\documentclass[lettersize,journal]{IEEEtran}
\usepackage{amsmath,amsfonts}
\usepackage{algorithmic}
\usepackage{algorithm}
\usepackage{array}
\usepackage[caption=false,font=normalsize,labelfont=sf,textfont=sf]{subfig}
\usepackage{textcomp}
\usepackage{stfloats}
\usepackage{url}
\usepackage{verbatim}
\usepackage{graphicx}
\usepackage{cite}
\usepackage{mathrsfs}
\usepackage{graphicx}   
\usepackage{booktabs}
\usepackage{threeparttable}
\usepackage{multirow}
\usepackage{bm}
\usepackage{color}
\newtheorem{Definition}{Definition}
\newtheorem{Assumption}{Assumption}
\usepackage{xcolor}         % colors

\definecolor{cGreen}{RGB}{0,150,0}%{128,255,128}
\definecolor{brown}{RGB}{139,64,0}

\definecolor{easy-top}{RGB}{160,208,208} %Mint Green
\definecolor{difficult-top}{RGB}{213,170,190} %Rose Pink

\newcommand{\meanstd}[2]{#1\,{\scriptsize$\pm$}\,#2}
\newcommand{\bestmeanstd}[2]{\textbf{#1}\,{\scriptsize$\pm$}\,#2}

\begin{document}

\title{Personalized Treatment Outcome Prediction from Scarce Data via Dual-Channel Knowledge Distillation and Adaptive Fusion}

\author{Wenjie Chen, Li Zhuang, Ziying Luo~\IEEEmembership{Student Member,~IEEE}, Yu Liu, Jiahao Wu, Shengcai Liu~\IEEEmembership{Member,~IEEE}

\thanks{
	Manuscript received (date to be filled in by Editor). This work was supported by National Natural Science Foundation of China (Grant No. 72401105, 62206051, and 72504104) and  by  “Zhishan” Scholars Programs of Southeast University (Grant No. 2242023R40056). \textit{(Corresponding Author: Shengcai Liu)}
	
	Wenjie Chen and Yu Liu are with the School of Information Management, Central China Normal University, Wuhan 430079, China. (email: chenwj6@ccnu.edu.cn; liuyuforwh@ccnu.edu.cn)
	
	Li Zhuang is with the School of Cyber Science and Engineering, Southeast University, Nanjing 211189, China. (email: lizhuanglily@gmail.com)
	
	Ziying Luo, Jiahao Wu, and Shengcai Liu are with the Guangdong Provincial Key Laboratory of Brain-inspired Intelligent Computation, Department of Computer Science and Engineering, Southern University of Science and Technology, Shenzhen 518055, China. (email: luozy2025@mail.sustech.edu.cn; jiahao.wu@connect.polyu.hk; liusc3@sustech.edu.cn)

}

}

% The paper headers
\markboth{Journal of \LaTeX\ Class Files,~Vol.~XX, No.~X, XX~XX}%
{Chen \MakeLowercase{\textit{et al.}}: Personalized Treatment Prediction from Scarce Trial Data via Dual-Channel Knowledge Distillation and Adaptive Fusion}

\maketitle

\begin{abstract}
Personalized treatment outcome prediction based on trial data for small-sample and rare patient groups is a critical task in precision medicine. However, the high cost and scarcity of trial data limit the prediction performance. To address this issue, we propose a cross-fidelity knowledge distillation and adaptive fusion network (CFKD-AFN), which leverages abundant but low-fidelity simulation data to enhance the prediction on scarce but high-fidelity trial data. 
CFKD-AFN incorporates a dual-channel knowledge distillation module to extract complementary knowledge from the low-fidelity model, along with an attention-guided fusion module to adaptively integrate multi-source information.
Experiments on chronic obstructive pulmonary disease show that CFKD-AFN reduces the mean squared error by 6.67\%–74.55\% and the mean absolute percentage error by 1.43\%–51.54\% compared to the evaluated competing methods and remains robust to varying high-fidelity dataset sizes.
Furthermore, we extend the CFKD-AFN framework to an interpretable variant {\color{black}for exploratory analysis of feature-attribution patterns associated with treatment outcomes.}
\end{abstract}


\section{Introduction}

\IEEEPARstart{P}{ersonalized} medicine is a healthcare paradigm that tailors optimal treatment strategies to individual patients based on variations in genes, environment, and lifestyle \cite{wang2023precision}.
With its potential to enhance healthcare quality and operational efficiency, it has become an emerging and rapidly evolving area of research in computational intelligence \cite{liu2024semi, alvi2022long, yang2021robust}.
One of the core computational challenges in personalized medicine is personalized treatment outcome prediction (PTOP), which {\color{black}is a treatment-specific prediction task and} aims to predict the potential efficacy of a specific therapeutic regimen for a given patient based on individual differences.
Driven by advances in machine learning and bioinformatics, PTOP is increasingly being applied to more diverse clinical prediction settings~\cite{kuenzi2020predicting, simon2022interpretable}.

Current approaches for PTOP can be classified into (but not limited to) retrospective studies based on historical data (RSHD) \cite{zhang2020errors,kuenzi2020predicting, ma2023kgml, simon2022interpretable}  and prospective studies based on trial data (PSTD) \cite{yao2019prediction, koesmahargyo2020accuracy, su2018random,grzenda2021machine}.
RSHD constructs predictive models by analyzing historical patient records, such as electronic health records, genomic data, medical images, and other clinical databases.
For example, Turki et al. \cite{turki2019clinical} utilized a public genetic database to predict the response of specific cancer types to clinical drugs.
With the rapid development of healthcare information technology, healthcare institutions have collected vast amounts of medical data, facilitating the wide adoption of RSHD in PTOP research.
Such methods are suitable for predicting treatment outcomes in large patient populations and routine cases, but are less effective for rare patient groups\footnote{Rare patient groups refer to those lacking sufficient high-fidelity trial data, either due to disease rarity or difficulties in clinical enrollment.}.
The reasons are twofold.
First, historical data usually originate from routine clinical practice, where the collection process is often loosely controlled.
Factors such as data collection methods and clinicians' implicit biases can introduce inherent biases into the data \cite{perets2024inherent}.
This makes it difficult for the prediction model to learn the subtle characteristics of rare cases.
Second, retrospective data lacks counterfactual information, meaning that the model can only infer the results of implemented interventions and cannot truly assess the possible efficacy of alternative options for specific individuals~\cite{wang2024rcfr}.
This severely limits the predictive accuracy and generalization capabilities for rare patient groups.

\begin{figure*}[!t]
	\centering
	\includegraphics[width=0.98\textwidth]{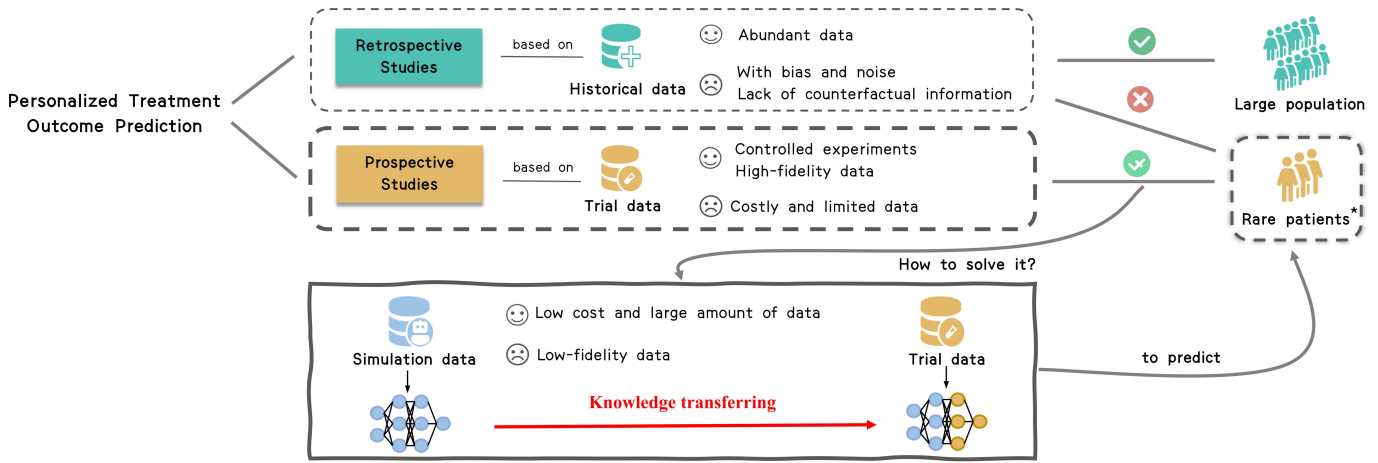} 
	\caption{Illustration of PTOP (RSHD vs. PSTD), the challenge of scarce high-fidelity data in PSTD, and the idea of using simulation to tackle this challenge.}
	\label{fig: Introduction} 
\end{figure*}

The other approach for PTOP, PSTD, leverages rigorously designed trials, including N-of-1, intervention-matching, and adaptive designs \cite{goetz2018personalized}, enabling the accurate assessment of new therapies in randomized and well-controlled environments.
Compared to RSHD, PSTD offers more advantages for small samples or rare groups, such as evaluating new drugs and treatment protocols. 
Nevertheless, PSTD is expensive.
Long recruitment periods and high financial costs make it difficult to collect sufficient patient records, while ethical and safety concerns impose additional constraints.
These limitations restrict the scalability and real-time applications of prospective studies.
For example, it is difficult to implement clinical trials  for patients with complex disease progression or those at high risk from experimental drugs.
Consequently, developing methods that accurately predict treatment outcome based on costly clinical data for rare patient groups has become a significant research problem in PTOP.
Fig.~\ref{fig: Introduction} illustrates the distinction between the RSHD and PSTD approaches, and highlights the research problem studied in this paper.

To tackle the challenge of data scarcity noted above, simulation has emerged as a practical tool for generating large volumes of low-cost data, with potential to enhance PTOP based on costly trial data. 
Approaches such as microsimulations~\cite{ayer2016heterogeneity} integrate empirical data and expert knowledge to model patient responses.
However, simulation inevitably oversimplifies real-world complexity, producing lower-fidelity data compared with clinical trials.
Multi-fidelity transfer learning is a common approach to leverage multi-fidelity data for modeling.
It transfers knowledge from abundant low-fidelity data to tasks with limited high-fidelity data to improve model performance, and has been successfully applied in fields such as materials science \cite{ghane2025multi}, engineering \cite{yue2024improving}, and drug discovery \cite{buterez2024transfer}.
This approach also offers us a promising framework for integrating high-fidelity clinical data with large-scale low-fidelity simulation data to improve PTOP.
% Figure~\ref{fig: Introduction} shows the research focus of this paper and our solution approach.

Existing multi-fidelity transfer learning methods can be broadly categorized into two types: pretraining-fine-tuning methods \cite{tang2025multi, zhang2024multi, de2020transfer, kalimullah2023probabilistic, lyu2023multi}, and multi-level data fusion methods \cite{meng2020composite, guo2022multi}.
The former trains a model on abundant low-fidelity data and then fine-tunes it using high-fidelity data.
The latter also trains a model on low-fidelity data, but combines its predicted output with high-fidelity data to train a separate network for high-fidelity outputs.
However, directly applying these methods to our PTOP problem is problematic, primarily due to two challenges.
The first is the extreme scarcity of high-fidelity data.
That is, clinical data (high-fidelity) is often far smaller than simulation data.
When high-fidelity samples are extremely limited (e.g., only ten samples), the pretraining-fine-tuning approach easily leads to overfitting and ineffective transfer.
The second is the distribution discrepancy.
That is, simulation models inherently simplify real-world scenarios, causing the source (low-fidelity) and target (high-fidelity) domains to differ in their marginal feature distributions and conditional outcome distributions.
The multi-level data fusion methods, by only using the predicted outputs as distilled knowledge, ignore these crucial feature-level differences.
This makes them unable to address the distribution discrepancy, and even introduces noise and misguides the training process.

To address these limitations, this paper proposes a cross-fidelity knowledge distillation and adaptive fusion network (CFKD-AFN).
To mitigate data scarcity, CFKD-AFN leverages a dual-channel knowledge distillation module to fully exploit the rich information in low-fidelity data, ensuring stable training even with very limited high-fidelity samples.
To handle distribution differences, it employs an attention-guided adaptive fusion module combined with multi-level knowledge transfer to avoid noisy transfer and enable smoother cross-domain knowledge integration.
Specifically, the dual-channel knowledge distillation module extracts complementary knowledge from the low-fidelity model: macroscopic prediction knowledge via its predicted outputs, and microscopic representational knowledge via  its high-dimensional feature representations. 
Subsequently, the attention-guided fusion module adaptively fuses high-fidelity inputs with the dual knowledge using a weight allocation module. 
Finally, a dedicated network converts the fused features into the final prediction. 
The main contributions of this paper are three-fold, as described below.
\begin{enumerate}
	\item This paper addresses the critical challenge of predicting personalized treatment outcome for small samples or rare groups using limited and costly clinical data. To tackle this, simulation techniques are used to enhance prediction performance by transferring knowledge from abundant low-fidelity simulation data to limited high-fidelity clinical data. 
    A novel multi-fidelity transfer learning method, CFKD-AFN, is proposed, which integrates a dual-channel knowledge distillation module to capture both macroscopic prediction knowledge and microscopic representational knowledge, and an attention-guided fusion module to adaptively fuse information from multiple sources.
	\item Experiments on treatment outcome prediction for the chronic obstructive pulmonary disease (COPD) show that CFKD-AFN improves prediction accuracy by 6.67\% to 74.55\% compared to the evaluated competing methods.
    It also exhibits strong robustness across high-fidelity datasets of different sizes, and performs especially well in extremely small high-fidelity sample scenarios---even when the sample size is as small as ten.
    \item  An interpretable extension (iCFKD-AFN) is also developed that employs mutual information-based disentangled representation learning to {\color{black}explore partially separated feature-attribution patterns associated with treatment outcomes.}
\end{enumerate}

The remainder of this paper is organized as follows.
Section~\ref{Literature Review} reviews related work on PTOP and multi-fidelity transfer learning.
Section~\ref{Problem Formulation and Algorithm} formalizes the multi-fidelity transfer learning problem for PTOP and details the proposed CFKD-AFN.
In Section~\ref{Case Study}, a comprehensive experimental evaluation on COPD is presented, along with the interpretable extension iCFKD-AFN in Section~\ref{Exploration of Model Interpretability}.
Finally, Section~\ref{Conclusion} concludes the paper and outlines potential directions for future work.

\section{Literature Review}\label{Literature Review}

\subsection{Review of PTOP}
\label{Review of personalized treatment outcome prediction}
As mentioned before, retrospective studies based on historical data and prospective studies based on trial data are the two main approaches in PTOP.
RSHD leverages various types of historical data to build predictive models.
The availability of large-scale databases enables these studies to uncover statistical patterns effectively.
For example, Kuenzi et al.~\cite{kuenzi2020predicting} developed an interpretable deep learning model called DrugCell, which was trained on a dataset of tumor cell lines’ responses to various drugs.
The model can predict drug efficacy and uncover the underlying biological mechanisms.
Simon et al.~\cite{simon2022interpretable} used electronic medical records to compare the effectiveness of deep learning models and interpretable models based on cluster analysis in predicting the risk of drug-induced long-QT syndrome.
Zhang et al.~\cite{zhang2020errors} used time series data from electronic health records to model treatment-response curves by combining parametric response functions, individual sharing information, and the Gaussian process. 
Ma et al.~\cite{ma2023kgml} constructed a KGML-xDTD framework, which used an open-source biomedical knowledge graph to predict the treatment probability of drugs and diseases and explain the prediction results.  

Unlike RSHD, PSTD utilizes rigorously designed clinical trial data to predict the outcome and is suitable for small samples or rare patient groups, such as in assessing the clinical applicability of new drugs and therapies.
Yao et al.~\cite{yao2019prediction} selected traditional machine learning classifiers,  such as decision trees, and conducted a follow-up experiment on 287 patients to predict the outcomes of newly diagnosed epilepsy patients in the antiepileptic drug treatment.
Koesmahargyo et al.~\cite{koesmahargyo2020accuracy} used the XGBoost model to dynamically predict medication adherence risk by remotely measuring drug dosage in real time.
Su et al.~\cite{su2018random} proposed a random forest method based on interactive trees to estimate individual treatment effects using the acupuncture headache trial data.
Grzenda et al.~\cite{grzenda2021machine} collected clinical trials conducted with depressed adults and leveraged classical machine learning methods to predict depression treatment outcomes.

Beyond RSHD and PSTD, simulation methods model patients’ physiology, pathology, and treatment responses in a cost-effective way, and are widely used to support personalized treatment decision-making, especially when real-world trial data is scarce or costly to obtain.
For example, Du et al.~\cite{du2024contextual} proposed an efficient ranking-selection algorithm for identifying personalized treatment plans in simulation-based personalized medicine, which could distinguish the optimum from suboptimal options across diverse patient populations.
Doudican et al.~\cite{doudican2015personalization} applied the cancer physiology simulation technology to screen various molecular-targeted drugs across different myeloma patient profiles to identify effective and synergistic treatment strategies. 

While clinical trial–based methods are advantageous for predicting treatment outcome in small-sample or rare-population scenarios, their development is hindered by limited sample sizes and high costs.
To address these problems, this paper proposes CFKD-AFN that integrates small-scale high-fidelity clinical data with abundant low-fidelity simulation data to enhance PTOP performance.

\subsection{Review of Multi-fidelity Transfer Learning}
\label{Review of multi-fidelity transfer learning}
Multi-fidelity transfer learning is a transfer learning paradigm designed to handle data from multiple fidelity levels.
Two commonly used approaches are the pretraining–fine-tuning approach and the multi-level data fusion approach.
However, naive pretraining–fine-tuning methods often perform poorly when high-fidelity data are scarce and there are significant discrepancies between high- and low-fidelity data.
Several studies have attempted to address these issues by enhancing the transfer and fine-tuning mechanisms.
For example, Buterez et al.~\cite{buterez2024transfer} addressed multi-fidelity transfer learning for graph neural networks.
They enhanced the readout transferability of graph neural networks by introducing a supervised variational graph autoencoder to learn a structured chemical latent space. 
Zhang et al.~\cite{zhang2024multi} proposed the MF-TLNN framework, which integrated an autoencoder with traditional pretraining–fine-tuning strategies to capture the relationships between low- and high-fidelity models.
De et al.~\cite{de2020transfer} proposed a bi-fidelity weighted learning method, where a low-fidelity network was pre-trained on low-fidelity data and refined using synthetic data from a Gaussian process model trained on high-fidelity data, with learning rates guided by the Gaussian process.  

Multi-level data fusion methods also begin by training a network on low-fidelity data.
The predictions generated by the low-fidelity network are then combined with high-fidelity data and input into a new network to produce the final prediction.
Building on this framework, researchers have explored various strategies for effectively combining these data sources.
Meng et al.~\cite{meng2020composite} proposed a composite neural network with three parts.
The first network, trained on low-fidelity data, is connected to two high-fidelity networks 
to capture both nonlinear and linear relationships between low- and high-fidelity data.
Guo et al.~\cite{guo2022multi} introduced a multi-layer learning method with two- and three-step versions. 
In the three-step model, a network is first trained on low-fidelity data, and its predictions are linearly combined with high-fidelity inputs.
This fused output, along with the predicted values and high-fidelity data, is used to train the high-fidelity prediction model.

Unlike existing methods, CFKD-AFN transfers both output-level and representation-level knowledge from low-fidelity data and fuses them with the high-fidelity input. Their relative contributions are learned jointly with the high-fidelity predictor, enabling more selective transfer under cross-fidelity distribution discrepancies.

\begin{table*}[htbp]
	\centering
	\caption{Summary of the notations}
	\begin{tabular}{ccc}
		\toprule
		\textbf{Notation}  & \textbf{Meaning} \\
		\midrule
		$n$  & Dimension of input features \\
		\midrule
		$\textbf{x}_l=\{x_l^1,...,x_l^n\}$  & Low-fidelity input \\
		\midrule
		$\textbf{x}_h=\{x_h^1,...,x_h^n\}$  & High-fidelity input \\
		\midrule
		$y_l$  & Low-fidelity label \\
		\midrule
		$y_h$  & High-fidelity label \\
		\midrule
		$\hat{y}_l$  & Predicted low-fidelity label \\
		\midrule
		$\hat{y}_h$  & Predicted high-fidelity label \\
		\midrule
		$y_{lh}$  & Predicted output by the low-fidelity model with high-fidelity data  $\textbf{x}_h$ as the input \\
		\midrule
		${\textbf{y}}_{feature}$  & Last-hidden-layer representation produced by the frozen low-fidelity model for the high-fidelity input \(\mathbf{x}_h\) \\
		\midrule
		${\textbf{F}}_{fused}$  & $3D$-dimensional representation produced by the attention-guided fusion module \\
		\bottomrule
	\end{tabular}
	\label{tab: notations}
\end{table*}

\section{Problem Formulation and Proposed Method}
\label{Problem Formulation and Algorithm}
This section first formulates the multi-fidelity transfer learning problem for PTOP, and then presents the CFKD-AFN method.

\subsection{Problem Formulation}
PTOP based on trial data aims to learn a prediction function $f(\cdot): \textbf{x}\rightarrow y$  where $\textbf{x}=\{x_1,...,x_n\}$ denotes the $n$-dimensional patient features (e.g., gender and age) and $y$ represents the treatment outcome (e.g., quality-adjusted life years).
In practice, the trial data are costly and limited, which constrains the generalization of the predictive model.
To this end, we leverage abundant simulation data to augment trial data and introduce multi-fidelity transfer learning to enhance predictive performance.
Consequently, the original prediction problem is reformulated as a multi-fidelity transfer learning task for PTOP. 

Without loss of generality, let a domain $\mathcal{D}=\{\mathcal{X}, P(\textbf{x})\}$ where $\mathcal{X}$ is the feature space, $P(\textbf{x})$ denotes the marginal probability distribution, and $\textbf{x}=\{x_1,...,x_n\}\in\mathcal{X}$.
Let a task $\mathcal{T}=\{\mathcal{Y}, P(y|\textbf{x})\}$ where $\mathcal{Y}$ represents the label space, $P(y|\textbf{x})$ is the conditional label distribution, and $y\in\mathcal{Y}$. 
Then, the multi-fidelity transfer learning problem for PTOP is formulated and two assumptions are made as follows. 

\begin{Definition}
	Given a source domain $\mathcal{D}_l$ with task $\mathcal{T}_l$ (low-fidelity simulation data) and a target domain $\mathcal{D}_h$ with task $\mathcal{T}_h$ (high-fidelity trial data), multi-fidelity transfer learning aims to learn a target predictive function $f_h(\cdot)$ in $\mathcal{D}_h$ by exploiting knowledge from $\mathcal{D}_l$ and $\mathcal{T}_l$.
\end{Definition}

\begin{Assumption}
	The source and target domains share the same feature and label spaces while their marginal and conditional distributions differ, that is,  $P_l(\textbf{x}) \neq P_h(\textbf{x})$, $P_l(y|\textbf{x}) \neq P_h(y|\textbf{x})$. 
\end{Assumption}
\begin{Assumption}
	Let \(t_l\) and \(t_h\) denote the numbers of low- and high-fidelity samples, respectively. We assume that $0<t_h\ll t_l$.
\end{Assumption}

These two assumptions are reasonable and commonly satisfied in reality. For Assumption 1, simulation and trial data may share the same patient features and outcome definition, while differences in population sampling and simulation fidelity result in different feature and conditional outcome distributions. Assumption 2 is consistent with our research setting, where abundant low-fidelity simulation data are used to improve PSTD-based PTOP with scarce high-fidelity trial samples.

\subsection{The Proposed CFKD-AFN Method}
We first introduce the basic architectures of previous algorithms: pretraining-fine-tuning and multi-level data fusion methods, and then detail the CFKD-AFN method.
Table~\ref{tab: notations} summarizes the notations used here.

\subsubsection{Traditional Algorithms}
\label{Section of Traditional Algorithms}
\begin{figure*}[!t]
	\centering
	\includegraphics[width=0.7\textwidth]{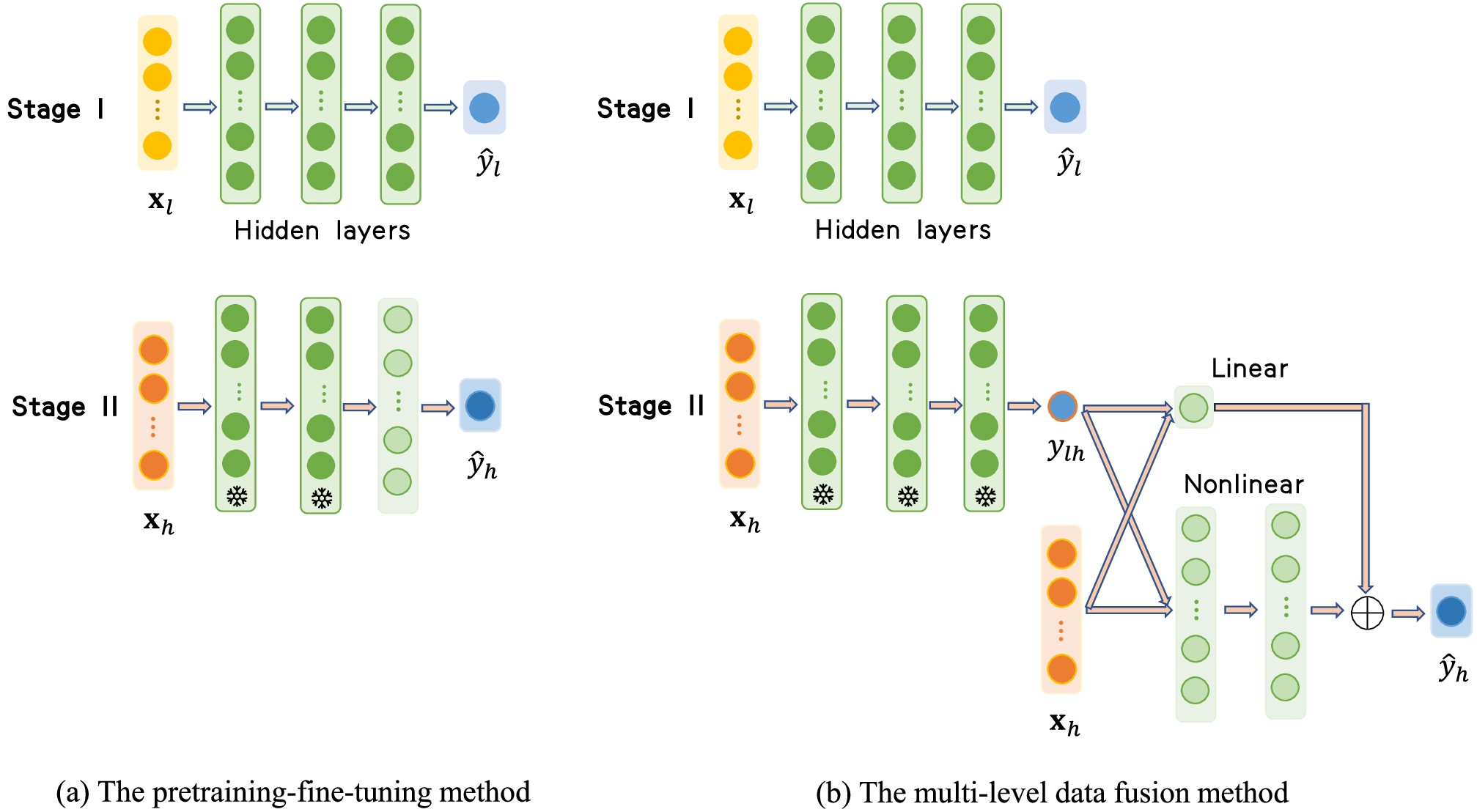}%
	\caption{The architectures of the pretraining-fine-tuning and the multi-level data fusion methods.}
	\label{fig: Traditional algorithms}
\end{figure*}

Pretraining-fine-tuning methods and multi-level data fusion methods both include two stages.
The first stage trains a low-fidelity network based on the low-fidelity data $(\textbf{x}_l, y_l)$.
In the second stage, the pretraining–fine-tuning methods freeze the parameters of the front part of the network, and the remaining part is fine-tuned using high-fidelity data $(\textbf{x}_h, y_h)$ as shown in Fig.~\ref{fig: Traditional algorithms}(a).
This strategy facilitates adaptation to the target domain while preserving transferable knowledge acquired from the low-fidelity source domain. 

Unlike the approaches that fix the initial layers to preserve source domain knowledge, the multi-level data fusion method transfers knowledge via $y_{lh}$ as an intermediate representation.
Fig.~\ref{fig: Traditional algorithms}(b) illustrates an example.
It directly concatenates $y_{lh}$ with $\textbf{x}_h$, then processes them through both a linear branch $f_{\text{lin}}(\cdot)$ and a nonlinear branch $f_{\text{nonlin}}(\cdot)$.
The outputs $f_{\text{lin}}(\textbf{x}_h,y_{lh})$ and $f_{\text{nonlin}}(\textbf{x}_h,y_{lh})$ are then combined to produce the final prediction $\hat{y}_h$.

\begin{figure*}[!t]
	\centering
	\includegraphics[width=0.7\textwidth]{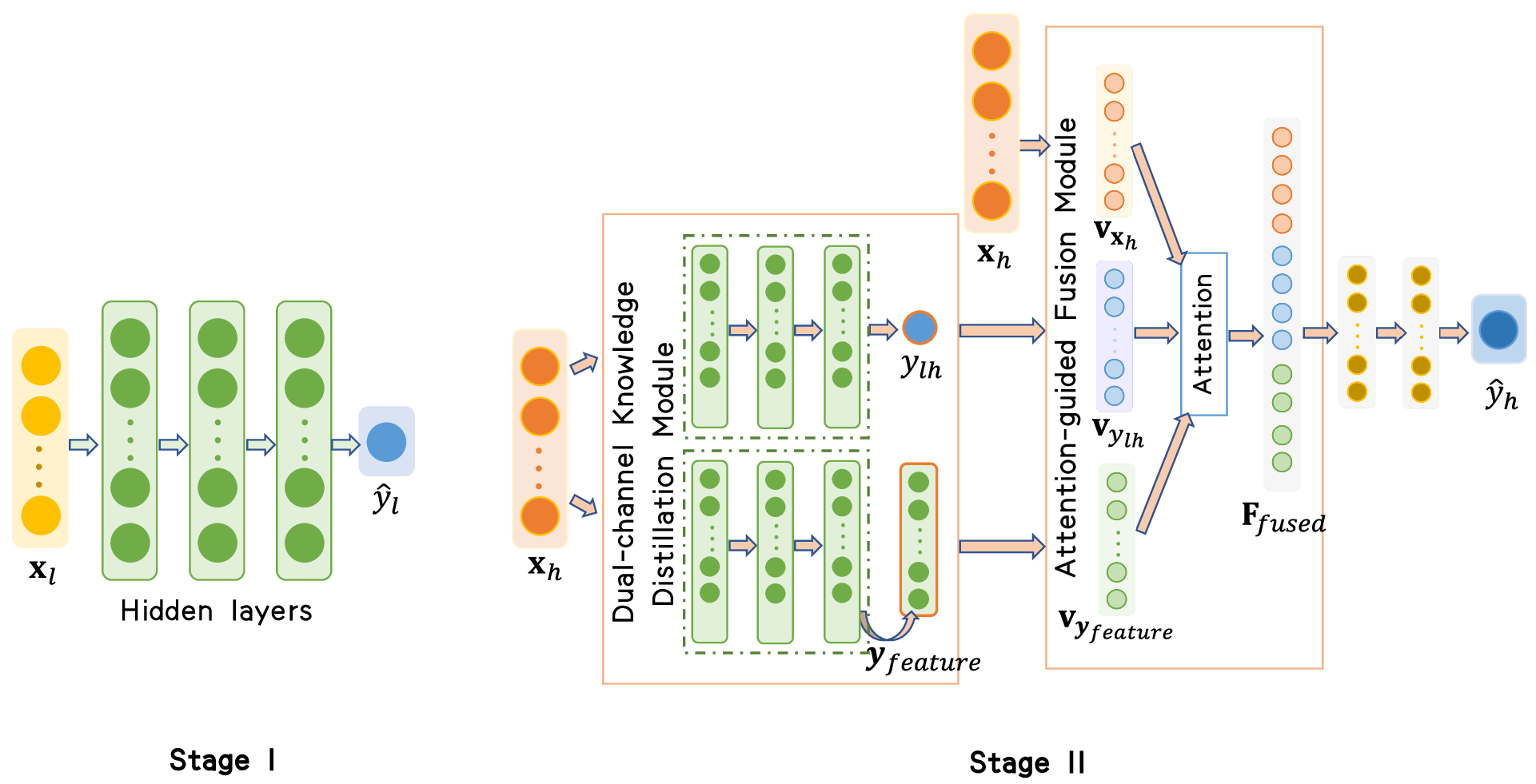} 
	\caption{The architecture of CFKD-AFN.}
	\label{fig: CFKD-AFN framework} 
\end{figure*}

\begin{algorithm}[!t]
	\caption{Cross-Fidelity Knowledge Distillation and Adaptive Fusion Network (CFKD-AFN)}
	\begin{algorithmic}[1]
		\REQUIRE Low-fidelity dataset $\mathcal{S}_l=\{(\mathbf{x}_{l,i}, y_{l,i})\}_{i=1}^{t_l}$, high-fidelity dataset $\mathcal{S}_h=\{(\mathbf{x}_{h,i}, y_{h,i})\}_{i=1}^{t_h}$, a new high-fidelity input $\mathbf{x}_{h}^*$
		\ENSURE Predicted high-fidelity output $\hat{y}_h^*$
		
		\item[] \textbf{Stage I: Training a low-fidelity model $f_l$}
		\STATE Use the low-fidelity dataset $\mathcal{S}_l$ to train a low-fidelity model $f_l$ based on the loss function $Loss = (1/{t_l})\cdot \sum_{i=1}^{t_l} \left( y_{l,i} - \hat{y}_{l,i} \right)^2$;
		
		\item[] \textbf{Stage II: Training a high-fidelity model $f_h$}
		\STATE Initialize the parameters $\bm{\theta}_h$ of $f_h$ and freeze $f_l$;
		\WHILE{the termination condition is not met}
		\FOR{%each high-fidelity sample $(\mathbf{x}_{h}, y_{h})\in S_h$
		$i=1,\ldots,t_h$}
		\STATE  Feed $\mathbf{x}_{h,i}$ into $f_l$;
		\STATE Extract macroscopic knowledge: $y_{lh,i} = f_l(\mathbf{x}_{h,i})$;
		\STATE Extract microscopic knowledge: $\mathbf{y}_{feature,i}$ from the last hidden layer of $f_l$;
		\STATE Apply linear transformation to $\mathbf{x}_{h,i}$, $y_{lh,i}$, and $\mathbf{y}_{feature,i}$ separately to obtain the aligned vectors  $\mathbf{v}_{\mathbf{x}_{h,i}}$, $\mathbf{v}_{y_{lh,i}}$, and $\mathbf{v}_{\mathbf{y}_{feature,i}}$;
		\STATE Compute the weights and concatenate the weighted vectors to obtain the fused representation ${\textbf{F}}_{fused,i}$;
		\STATE Pass \(\mathbf F_{\mathrm{fused},i}\) through the prediction layers of \(f_h\) to obtain $\hat{y}_{h,i}$;
		\ENDFOR
		\STATE Update the parameters $\bm{\theta}_h$ by the gradient-based optimizer based on the loss function $Loss = (1/{t_h})\cdot \sum_{i=1}^{t_h} \left( y_{h,i} - \hat{y}_{h,i} \right)^2$;
		\ENDWHILE
		\item[] \textbf{Stage III: Predicting the high-fidelity output $\hat{y}_h^*$}
		\STATE Given a new high-fidelity input $\mathbf{x}_{h}^*$, use $f_l$ and $f_h$ to predict $\hat{y}_h^*$.
		\RETURN $\hat{y}_h^*$
	\end{algorithmic}
    \label{Algorithm 1}
\end{algorithm}

\subsubsection{CFKD-AFN}
Unlike the previous approaches that rely on single-level knowledge or static fusion strategies, CFKD-AFN fully captures the low-fidelity knowledge and adaptively integrates information from low- and high-fidelity domains.
As illustrated in Fig.~\ref{fig: CFKD-AFN framework}, CFKD-AFN consists of two main components:
(i) a dual-channel knowledge distillation module to extract macroscopic and microscopic knowledge from the low-fidelity source domain, and
(ii) an attention-guided fusion module to adaptively fuse multi-source information.
These two modules are detailed as follows. 

\begin{itemize}
	\item \textit{Dual-channel knowledge distillation module.}
	In this module, the pretrained low-fidelity model \(f_l\) is frozen and used as a knowledge provider for the high-fidelity inputs. Each high-fidelity input \(\mathbf{x}_h\) is fed into \(f_l\) to obtain the output-level signal \(y_{lh}\) and the representation-level signal \(\mathbf{y}_{{feature}}\). These two signals serve as auxiliary information for the high-fidelity predictor.
    Specifically, $y_{lh}$ provides macroscopic  knowledge and offers a reliable reference for the output. The representation ${\textbf{y}}_{feature}$ extracted from the last hidden layer is used to capture microscopic knowledge because it is most directly optimized for the prediction task and avoids additional layer-alignment.
    By jointly exploiting the complementary knowledge, CFKD-AFN helps reduce overfitting when high-fidelity data are scarce and offers multi-level inputs to enhance cross-fidelity knowledge fusion.
	\item \textit{Attention-guided fusion module.} Adaptive mechanisms enable a model to adjust the contributions of different components \cite{11601056}. Following this principle, our attention-guided fusion module learns the weights for three information sources. Specifically, this module takes the prediction $y_{lh}$, the high-dimensional feature representation ${\textbf{y}}_{feature}$, and the high-fidelity data $\textbf{x}_h$ as inputs.
    Each input is first linearly transformed as follows.
	\begin{equation*}
		\begin{aligned}
			\mathbf{v}_{y_{lh}} &= \mathcal{L}_{y_{lh}}(y_{lh}),\\
			\mathbf{v}_{{\textbf{y}}_{feature}} &= \mathcal{L}_{{\textbf{y}}_{feature}}({\textbf{y}}_{feature}) \\
			\mathbf{v}_{\textbf{x}_h} &= \mathcal{L}_{\textbf{x}_h}(\textbf{x}_h)
		\end{aligned},
		\label{eq:feature_fusion}
	\end{equation*}
	where $\mathcal{L}_{y_{lh}}: \mathbb{R} \to \mathbb{R}^D$, $\mathcal{L}_{{\textbf{y}}_{feature}}: \mathbb{R}^{D_{{\textbf{y}}_{feature}} } \to \mathbb{R}^D$, and $\mathcal{L}_{\textbf{x}_h}: \mathbb{R}^n  \to \mathbb{R}^D$ are linear transformation functions. Here, $D_{{\textbf{y}}_{feature}}$ and $D$ are the dimensions of ${\textbf{y}}_{feature}$ and the transformed representation, respectively.  Then, an adaptive weighted attention mechanism combines them into a fused feature representation ${\textbf{F}}_{fused}$ as follows.
	\begin{equation*}
		\begin{aligned}
			\alpha_k &= \frac{\exp(w_k)}{\sum_{j=1}^3 \exp(w_j)}, \quad k \in \{1,2,3\},  \\
			{\textbf{F}}_{fused} &= \text{Concat}\left( \alpha_1 \cdot \mathbf{v}_{y_{lh}},\ \alpha_2 \cdot \mathbf{v}_{{\textbf{y}}_{feature}},\ \alpha_3 \cdot \mathbf{v}_{\textbf{x}_h} \right)
		\end{aligned}
		\label{eq:feature_fusion}
	\end{equation*} where $\mathbf{w} = [w_1, w_2, w_3] \in \mathbb{R}^3$ represents the attention weight vector, $\alpha_i (i=1,2,3)$ denotes the attention score, and $\text{Concat}(\cdot)$ is the concatenation operation along the channel dimension. {\color{black}The attention weights are trainable parameters shared across all samples within a treatment task.}
	This design utilizes complementary priors from the low-fidelity model while preserving essential high-fidelity information.
    The linear transformation aligns the heterogeneous inputs in a common latent space, and the adaptive attention mechanism balances their contributions, preventing noisy features from dominating the fusion.
    This allows CFKD-AFN to reduce transfer noise and achieve smoother integration of cross-domain knowledge.
\end{itemize}

Algorithm~\ref{Algorithm 1} presents the details of CFKD-AFN. 
In Stage I, CFKD-AFN trains the low-fidelity model $f_l$ using low-fidelity dataset $S_l$ based on the loss function $Loss = (1/{t_l})\cdot \sum_{i=1}^{t_l} \left( y_{l,i} - \hat{y}_{l,i} \right)^2$ (Line 1).
In Stage II, we initialize the parameters $\bm{\theta}_h$ of $f_h$, freeze  $f_l$  (Line 2) and then train $f_h$ (Lines 3-13). Specifically,
for each high-fidelity input $\textbf{x}_{h,i}(i=1,...,t_h)$, the dual-channel knowledge distillation module first uses the low-fidelity model $f_l$ to extract  the macroscopic knowledge $y_{lh,i}$ and microscopic knowledge ${\textbf{y}}_{feature,i}$ (Lines 5-7).
They are combined with $\textbf{x}_{h,i}$ and fed into the attention-guided fusion module.
Within this module, the inputs first pass through linear transformations to obtain the aligned vectors $\mathbf{v}_{\mathbf{x}_{h,i}}$, $\mathbf{v}_{y_{lh,i}}$, and $\mathbf{v}_{\mathbf{y}_{feature,i}}$ (Line 8).
An adaptive weighted attention mechanism then learns the relative importance of three vectors, and combines them to produce a fused representation ${\textbf{F}}_{fused,i}$ (Line 9).
Finally, ${\textbf{F}}_{fused,i}$ is passed through a predictor to generate the high-fidelity output $\hat{y}_{h,i}$ (Line 10).
The high-fidelity model is trained by the gradient-based optimizer based on the loss function $Loss = (1/{t_h})\cdot \sum_{i=1}^{t_h} \left( y_{h,i} - \hat{y}_{h,i} \right)^2$ (Line 12).
Stage II terminates when the condition is met.  For a new high-fidelity input \(x_h^*\), the frozen $f_l$ first produces \(y_{lh}^*\) and \(y_{{feature}}^*\), after which the trained \(f_h\) generates the prediction \(\hat y_h^*\) in Stage III (Line 14). CFKD-AFN returns the predicted high-fidelity output $\hat{y}_{h}^*$ (Line 15). Note that Algorithm 1 presents the full-batch optimization procedure for notational simplicity. The same procedure can also be implemented using mini-batch updates.

\section{Experimental Study on Treatment Outcome Prediction for COPD}
\label{Case Study}
This section evaluates CFKD-AFN on a simulation-based treatment outcome prediction task for chronic obstructive pulmonary disease (COPD).
The prediction accuracy of CFKD-AFN is compared with state-of-the-art methods across varying high-fidelity dataset sizes.
Ablation experiments are further conducted to evaluate the contribution of multi-source information to performance improvement.

\subsection{Application Background of COPD}
COPD is a common, preventable, and treatable respiratory disease that causes gradually worsening airflow limitation, shortness of breath, and cough.
With its global prevalence and significant medical burden, significant research efforts are focused on exploring new treatments to reduce healthcare costs, improve patients' quality of life, and support personalized medicine.

Given the limited availability of real-world trial data, we use a publicly available health economic simulation model~\cite{hoogendoorn2019broadening} to construct a controlled small-sample setting. The model incorporates patient characteristics, disease progression, and treatment effects to generate high-fidelity data as a proxy for trial data.
A low-fidelity simulation model is constructed by simplifying the original model.
For details on low-fidelity COPD modeling and high- and low-fidelity dataset generation, please refer to Appendix~\ref{app:Low-fidelity COPD Simulation Modeling} and Appendix~\ref{app:High- and Low-fidelity Data Generation}.

The COPD simulation model estimates quality-adjusted life-years (QALYs) of patients under different treatments based on fourteen characteristics---age, BMI, smoking status, number of pack-years, sex, heart failure, other cardiovascular disease, diabetes, depression, asthma/rhinitis, emphysema, concomitant conditions, eosinophil level, and bronchodilator responsiveness.
Four treatment options are tested in this study, as summarized in Table~\ref{tab: treatment options}. {\color{black}A separate predictive model is trained and evaluated for each treatment to predict the corresponding QALYs outcome.}

\begin{table}[tbp]
    \centering
    \caption{Details of the Four Treatments}
    \label{tab:treatment-options}
    \resizebox{\linewidth}{!}{
        \begin{tabular}{p{0.01\linewidth} p{0.34\linewidth} p{0.5\linewidth}}
            \toprule
            \textbf{No.} & \textbf{Treatment} & \textbf{Meaning} \\
            \midrule
            1 & Protect the pulmonary function 
              & Reduce the annual decline rate of the pulmonary function by 20\% \\
            \midrule
            2 & Delay the time of the acute exacerbation 
              & Extend the interval time of the acute exacerbation events by 30\% \\
            \midrule
            3 & Improve physical activity 
              & Increase the physical activity level by 4 points \\
            \midrule
            4 & Improve symptoms 
              & Reduce the probability of patients having dyspnea and cough symptoms by 20\% each \\
            \bottomrule
        \end{tabular}
    }
    \label{tab: treatment options}
    \vskip -0.15in
\end{table}
\subsection{Experimental Setting}
\subsubsection{Compared Methods}
Four representative approaches are selected: the high-fidelity method (HF), the pretraining–fine-tuning method (PFT), the multi-fidelity transfer learning neural network (MF-TLNN) \cite{zhang2024multi}, and the multi-fidelity data fusion model (MF-DF) \cite{zhang2024multi}.
HF and PFT serve as the baselines.
HF trains a neural network exclusively on high-fidelity data, whereas PFT follows the conventional pretraining–fine-tuning procedure described in Section~\ref{Section of Traditional Algorithms}.
MF-TLNN and MF-DF represent state-of-the-art pretraining–fine-tuning and multi-level data-fusion approaches, respectively.
In MF-TLNN, a low-fidelity neural network is trained alongside an autoencoder learned from $y_l$; both are fine-tuned with high-fidelity data, and their outputs are combined via a weighted sum to produce the final prediction.
The MF-DF method is detailed in Section~\ref{Section of Traditional Algorithms}.

\subsubsection{Performance Metric}
The methods are compared in terms of prediction accuracy.
Mean squared error (MSE) and mean absolute percentage error (MAPE) are chosen \cite{11481606}. MSE places greater weight on large absolute deviations, whereas MAPE reports the average relative error across patients with different QALYs magnitudes.

\subsubsection{Experimental Protocol}

For each treatment, 5000 low-fidelity samples and 1000 high-fidelity samples are generated.
The low-fidelity samples are used to train both the low-fidelity model and the autoencoder, with hyperparameters optimized via grid search. The autoencoder adopts the same hyperparameters as the low-fidelity model as suggested in~\cite{zhang2024multi}.
The high-fidelity samples are divided equally into two high-fidelity pools. The first pool is used for hyperparameter determination. Because the amount of high-fidelity data in practice is uncertain and the optimal configuration may vary with sample size, 20 subsets with sizes randomly drawn from the integer interval \([10,100]\) are sampled from the first pool. The hyperparameter configuration with the lowest average validation error on 20 experiments is selected for each method.
An early stopping strategy is applied to prevent overfitting during training.
Detailed hyperparameter setting procedures and results are provided in Appendix~\ref{app: Hyperparameters Setting}. The selected hyperparameters are then fixed for evaluation using the second pool, which will be described in the following subsections.

For fair comparison, all the experiments are conducted in Python and run on the same computing
platform, i.e., 12th Gen Intel (R) Core (TM) i7 - 12700 CPU with ~2.1GHz and 16GB of memory. 

\begin{table*}[!h]
	\centering
	\caption{Expected Performance Comparisons among Different Methods for Four Treatments}
	\resizebox{0.8\linewidth}{!}{
		\begin{threeparttable}
			\begin{tabular}%{cccccccc}
				{
					>{\centering\arraybackslash}m{2cm}
					>{\centering\arraybackslash}m{1.5cm} 
					>{\centering\arraybackslash}m{1.5cm} 
					>{\centering\arraybackslash}m{1.5cm} 
					>{\centering\arraybackslash}m{1.5cm} 
					>{\centering\arraybackslash}m{1.5cm} 
					>{\centering\arraybackslash}m{1.5cm} 
					>{\centering\arraybackslash}m{1.5cm}  
				}
				\toprule
				&                       &            & \textbf{HF}    & \textbf{PFT}   & \textbf{MF-TLNN} & \textbf{MF-DF} & \textbf{CFKD-AFN}       \\ 
				\midrule
				\multirow{4}{*}{Treatment 1} & \multirow{2}{*}{MSE}  & average       & 0.74  & 0.93  & 0.60     & 0.35  & \textbf{0.32} \\
				&                       & ratio (\%) & 56.76 & 65.59 & 46.67   & 8.57  & -             \\ \cmidrule{2-8} 
				& \multirow{2}{*}{MAPE} & average       & 12.19 & 14.55 & 12.36   & 8.24  & \textbf{8.08} \\
				&                       & ratio (\%) & 33.72 & 44.47 & 34.63   & 1.94  & -             \\ 
				\midrule
				\multirow{4}{*}{Treatment 2} & \multirow{2}{*}{MSE}  & average       & 0.54  & 0.49  & 0.38    & 0.20   & \textbf{0.16} \\
				&                       & ratio (\%) & 70.37 & 67.35 & 57.89   & 20.00    & -             \\ \cmidrule{2-8}  
				& \multirow{2}{*}{MAPE} & average       & 11.18 & 10.59 & 9.94    & 6.31  & \textbf{5.95} \\
				&                       & ratio (\%) & 46.78 & 43.81 & 40.14   & 5.71  & -             \\ 
				\midrule
				\multirow{4}{*}{Treatment 3} & \multirow{2}{*}{MSE}  & average       & 0.37  & 0.55  & 0.44    & 0.15  & \textbf{0.14} \\
				&                       & ratio (\%) & 62.16 & 74.55 & 68.18   & 6.67  & -             \\ \cmidrule{2-8}  
				& \multirow{2}{*}{MAPE} & average       & 9.35  & 11.37 & 10.95   & 5.59  & \textbf{5.51} \\
				&                       & ratio (\%) & 41.07 & 51.54 & 49.68   & 1.43  & -             \\ 
				\midrule
				\multirow{4}{*}{Treatment 4} & \multirow{2}{*}{MSE}  & average       & 0.49  & 0.44  & 0.43    & 0.18  & \textbf{0.16} \\
				&                       & ratio (\%) & 67.35 & 63.64 & 62.79   & 11.11 & -             \\ \cmidrule{2-8} 
				& \multirow{2}{*}{MAPE} & average       & 9.80   & 10.17 & 10.14   & 6.03  & \textbf{5.58} \\
				&                       & ratio (\%) & 43.06 & 45.13 & 44.97   & 7.46  & -             \\ 
				\bottomrule
			\end{tabular}
			\begin{tablenotes}
				\item[a] ``ratio'' indicates the relative improvement of CFKD-AFN over the competing method.
			\end{tablenotes}
		\end{threeparttable}
	}
	\label{tab: Expected Performance Comparison}
	\vskip -0.15in
\end{table*}

\subsection{Prediction Performance Comparison}
The CFKD-AFN method is evaluated against state-of-the-art approaches from two perspectives: expected performance and fine-grained performance, which assess the average predictive accuracy and the robustness of each algorithm across varying amounts of high-fidelity data, respectively. {\color{black}In Tables ~\ref{tab: Expected Performance Comparison}–\ref{tab: ablation experiments}, the relative improvement is calculated as \(\mathrm{ratio}=(E_m-E_{\mathrm{CFKD-AFN}})/E_m\times100\%\), where \(E_m\) and \(E_{\mathrm{CFKD-AFN}}\) denote the performance of competing method \(m\) and CFKD-AFN, respectively.}

\begin{table*}[tbp]
  \centering
  \caption{Fine-grained Performance Comparisons among Different Methods for Treatment 1}
  \resizebox{0.8\linewidth}{!}{
      \begin{threeparttable}
      \begin{tabular}{
          >{\centering\arraybackslash}m{2cm}
          >{\centering\arraybackslash}m{1.5cm}
          >{\centering\arraybackslash}m{2.2cm}
          >{\centering\arraybackslash}m{1.5cm}
          >{\centering\arraybackslash}m{1.5cm}
          >{\centering\arraybackslash}m{1.5cm}
          >{\centering\arraybackslash}m{1.5cm}
          >{\centering\arraybackslash}m{1.5cm}
        }
        \toprule
        & &  & \textbf{HF} & \textbf{PFT} & \textbf{MF-TLNN} & \textbf{MF-DF} & \textbf{CFKD-AFN} \\
        \midrule
    
        \multirow{4}{*}{10 samples} & \multirow{2}{*}{MSE}
          & average (± std) & \meanstd{2.35}{1.23} & \meanstd{1.70}{0.74} & \bestmeanstd{0.75}{0.27} & \meanstd{1.15}{0.41} & \meanstd{0.86}{0.37} \\
        & & ratio (\%)       & 63.40 & 49.41 & -14.67 & 25.22 & - \\ \cmidrule{2-8}
        & \multirow{2}{*}{MAPE}
          & average (± std) & \meanstd{24.07}{6.37} & \meanstd{20.50}{5.42} & \bestmeanstd{13.60}{2.91} & \meanstd{17.16}{3.81} & \meanstd{14.50}{4.04} \\
        & & ratio (\%)       & 39.76 & 29.27 & -6.62 & 15.50 & - \\
        \midrule
    
        \multirow{4}{*}{20 samples} & \multirow{2}{*}{MSE}
          & average (± std) & \meanstd{1.55}{0.71} & \meanstd{1.13}{0.39} & \meanstd{0.67}{0.25} & \meanstd{0.66}{0.32} & \bestmeanstd{0.55}{0.26} \\
        & & ratio (\%)       & 64.52 & 51.33 & 17.91 & 16.67 & - \\ \cmidrule{2-8}
        & \multirow{2}{*}{MAPE}
          & average (± std) & \meanstd{19.30}{3.92} & \meanstd{16.39}{2.74} & \meanstd{12.91}{2.69} & \meanstd{12.35}{2.39} & \bestmeanstd{11.20}{2.19} \\
        & & ratio (\%)       & 41.97 & 31.67 & 13.25 & 9.31 & - \\
        \midrule
    
        \multirow{4}{*}{40 samples} & \multirow{2}{*}{MSE}
          & average (± std) & \meanstd{0.59}{0.31} & \meanstd{0.78}{0.28} & \meanstd{0.61}{0.16} & \bestmeanstd{0.27}{0.15} & \meanstd{0.31}{0.12} \\
        & & ratio (\%)       & 47.46 & 60.26 & 49.18 & -14.81 & - \\ \cmidrule{2-8}
        & \multirow{2}{*}{MAPE}
          & average (± std) & \meanstd{11.98}{2.36} & \meanstd{13.67}{2.21} & \meanstd{12.51}{1.52} & \bestmeanstd{7.71}{1.97} & \meanstd{8.38}{1.51} \\
        & & ratio (\%)       & 30.05 & 38.70 & 33.01 & -8.69 & - \\
        \midrule
    
        \multirow{4}{*}{80 samples} & \multirow{2}{*}{MSE}
          & average (± std) & \meanstd{0.25}{0.06} & \meanstd{0.61}{0.11} & \meanstd{0.54}{0.08} & \meanstd{0.13}{0.03} & \bestmeanstd{0.12}{0.03} \\
        & & ratio (\%)       & 52.00 & 80.33 & 77.78 & 7.69 & - \\ \cmidrule{2-8}
        & \multirow{2}{*}{MAPE}
          & average (± std) & \meanstd{7.95}{1.02} & \meanstd{12.30}{1.29} & \meanstd{11.91}{1.08} & \meanstd{5.47}{0.65} & \bestmeanstd{5.20}{0.61} \\
        & & ratio (\%)       & 34.59 & 57.72 & 56.34 & 4.94 & - \\
        \midrule
    
        \multirow{4}{*}{100 samples} & \multirow{2}{*}{MSE}
          & average (± std) & \meanstd{0.19}{0.03} & \meanstd{0.58}{0.08} & \meanstd{0.54}{0.08} & \meanstd{0.11}{0.02} & \bestmeanstd{0.10}{0.02} \\
        & & ratio (\%)       & 47.37 & 82.76 & 81.48 & 9.09 & - \\ \cmidrule{2-8}
        & \multirow{2}{*}{MAPE}
          & average (± std) & \meanstd{6.96}{0.63} & \meanstd{12.09}{0.93} & \meanstd{11.82}{0.99} & \meanstd{5.04}{0.47} & \bestmeanstd{4.89}{0.43} \\
        & & ratio (\%)       & 29.74 & 59.55 & 58.63 & 2.98 & - \\
        \bottomrule
      \end{tabular}
      \begin{tablenotes}
        \item[a] ``ratio'' indicates the relative improvement of CFKD-AFN over the competing method.
      \end{tablenotes}
      \end{threeparttable}
    }
  \label{tab: Fine-grained Performance Comparison for T1}
\end{table*}

\begin{table*}[tbp]
  \centering
  \caption{Fine-grained Performance Comparisons among Different Methods for Treatment 2}
  \resizebox{0.8\linewidth}{!}{
      \begin{threeparttable}
        \begin{tabular}{
          >{\centering\arraybackslash}m{2cm}
          >{\centering\arraybackslash}m{1.5cm}
          >{\centering\arraybackslash}m{2.2cm}
          >{\centering\arraybackslash}m{1.5cm}
          >{\centering\arraybackslash}m{1.5cm}
          >{\centering\arraybackslash}m{1.5cm}
          >{\centering\arraybackslash}m{1.5cm}
          >{\centering\arraybackslash}m{1.5cm}
        }
          \toprule
          & & & \textbf{HF} & \textbf{PFT} & \textbf{MF-TLNN} & \textbf{MF-DF} & \textbf{CFKD-AFN} \\
          \midrule
    
          \multirow{4}{*}{10 samples} & \multirow{2}{*}{MSE}
            & average (± std) & \meanstd{2.30}{1.42} & \meanstd{1.16}{0.61} & \bestmeanstd{0.47}{0.36} & \meanstd{0.68}{0.52} & \meanstd{0.52}{0.39} \\
          & & ratio (\%) & 77.39 & 55.17 & -10.64 & 23.53 & - \\ \cmidrule{2-8}
          & \multirow{2}{*}{MAPE}
            & average (± std) & \meanstd{24.15}{8.16} & \meanstd{17.38}{4.60} & \bestmeanstd{10.73}{4.60} & \meanstd{12.68}{4.92} & \meanstd{11.22}{3.58} \\
          & & ratio (\%) & 53.54 & 35.44 & -4.57 & 11.51 & - \\
          \midrule
    
          \multirow{4}{*}{20 samples} & \multirow{2}{*}{MSE}
            & average (± std) & \meanstd{1.25}{0.53} & \meanstd{1.10}{0.63} & \meanstd{0.46}{0.17} & \meanstd{0.45}{0.14} & \bestmeanstd{0.32}{0.15} \\
          & & ratio (\%) & 74.40 & 70.91 & 30.43 & 28.89 & - \\ \cmidrule{2-8}
          & \multirow{2}{*}{MAPE}
            & average (± std) & \meanstd{18.07}{3.51} & \meanstd{15.89}{4.26} & \meanstd{11.56}{2.31} & \meanstd{10.90}{1.69} & \bestmeanstd{9.08}{2.08} \\
          & & ratio (\%) & 49.75 & 42.86 & 21.45 & 16.70 & - \\
          \midrule
    
          \multirow{4}{*}{40 samples} & \multirow{2}{*}{MSE}
            & average (± std) & \meanstd{0.63}{0.36} & \meanstd{0.47}{0.23} & \meanstd{0.42}{0.13} & \bestmeanstd{0.17}{0.07} & \bestmeanstd{0.17}{0.07} \\
          & & ratio (\%) & 73.02 & 63.83 & 59.52 & 0.00 & - \\ \cmidrule{2-8}
          & \multirow{2}{*}{MAPE}
            & average (± std) & \meanstd{11.90}{2.24} & \meanstd{10.61}{1.85} & \meanstd{10.44}{1.61} & \meanstd{6.33}{1.09} & \bestmeanstd{6.17}{1.19} \\
          & & ratio (\%) & 48.15 & 41.85 & 40.90 & 2.53 & - \\
          \midrule
    
          \multirow{4}{*}{80 samples} & \multirow{2}{*}{MSE}
            & average (± std) & \meanstd{0.20}{0.05} & \meanstd{0.34}{0.06} & \meanstd{0.34}{0.06} & \meanstd{0.07}{0.02} & \bestmeanstd{0.06}{0.02} \\
          & & ratio (\%) & 70.00 & 82.35 & 82.35 & 14.29 & - \\ \cmidrule{2-8}
          & \multirow{2}{*}{MAPE}
            & average (± std) & \meanstd{7.38}{0.94} & \meanstd{9.34}{0.97} & \meanstd{9.44}{0.84} & \meanstd{4.27}{0.44} & \bestmeanstd{3.99}{0.54} \\
          & & ratio (\%) & 45.93 & 57.28 & 57.73 & 6.56 & - \\
          \midrule
    
          \multirow{4}{*}{100 samples} & \multirow{2}{*}{MSE}
            & average (± std) & \meanstd{0.19}{0.06} & \meanstd{0.31}{0.05} & \meanstd{0.32}{0.05} & \meanstd{0.06}{0.02} & \bestmeanstd{0.05}{0.01} \\
          & & ratio (\%) & 73.68 & 83.87 & 84.38 & 16.67 & - \\ \cmidrule{2-8}
          & \multirow{2}{*}{MAPE}
            & average (± std) & \meanstd{7.09}{0.80} & \meanstd{9.17}{0.67} & \meanstd{9.31}{0.70} & \meanstd{3.95}{0.39} & \bestmeanstd{3.55}{0.38} \\
          & & ratio (\%) & 49.93 & 61.29 & 61.87 & 10.13 & - \\
          \bottomrule
        \end{tabular}
        \begin{tablenotes}
          \item[a] ``ratio'' indicates the relative improvement of CFKD-AFN over the competing method.
        \end{tablenotes}
      \end{threeparttable}
    }
  \label{tab: Fine-grained Performance Comparison for T2}
\end{table*}
\subsubsection{Expected Performance Comparison}
For expected performance, 50 repeated sampling trials are generated from the second high-fidelity pool, with their dataset sizes drawn from a uniform distribution over [10, 100].
The average predictive accuracy of each method is then assessed for each treatment.
Table~\ref{tab: Expected Performance Comparison} shows the expected performance comparison results for four treatments.
Based on these results, three observations can be made as follows.

\textbf{Observation 1: CFKD-AFN outperforms existing methods under varying amounts of high-fidelity data.}
CFKD-AFN consistently achieves superior predictive accuracy over the competing methods across all four treatments. Specifically, for Treatment 3, it yields the minimal average improvements of 6.67\% in MSE and 1.43\% in MAPE over MF-DF, while achieving the maximal improvements of 74.55\% in MSE and 51.54\% in MAPE over PFT.

\textbf{Observation 2: The conventional pretraining–fine-tuning method does not necessarily yield performance improvements.}
In Treatments 1 and 3, PFT performs worse than the HF approach.
This may be because traditional pretraining–fine-tuning methods are prone to negative transfer during migration.
When target samples are extremely limited, direct fine-tuning may lead to overfitting.

\textbf{Observation 3: MF-DF shows considerably better performance than the other three baseline methods.}
This may be attributed to its direct use of the one-dimensional prediction from the low-fidelity model as the input feature of the high-fidelity model, which effectively reduces information loss and noise interference.
However, such a straightforward approach fails to fully exploit the representations of the input data, particularly the high-dimensional features from the low-fidelity data.
When the amount of high-fidelity data is extremely small, the one-dimensional output of the low-fidelity model may provide insufficient information for transfer learning, potentially leading to performance degradation.
This conclusion will be further validated in the subsequent fine-grained analysis.

\begin{table*}[tbp]
  \centering
  \caption{Fine-grained Performance Comparisons among Different Methods for Treatment 3}
  \resizebox{0.8\linewidth}{!}{
      \begin{threeparttable}
        \begin{tabular}{
          >{\centering\arraybackslash}m{2cm}
          >{\centering\arraybackslash}m{1.5cm}
          >{\centering\arraybackslash}m{2.2cm}
          >{\centering\arraybackslash}m{1.5cm}
          >{\centering\arraybackslash}m{1.5cm}
          >{\centering\arraybackslash}m{1.5cm}
          >{\centering\arraybackslash}m{1.5cm}
          >{\centering\arraybackslash}m{1.5cm}
        }
          \toprule
          & & & \textbf{HF} & \textbf{PFT} & \textbf{MF-TLNN} & \textbf{MF-DF} & \textbf{CFKD-AFN} \\
          \midrule
    
          \multirow{4}{*}{10 samples} & \multirow{2}{*}{MSE}
            & average (± std) & \meanstd{1.87}{0.77} & \meanstd{4.16}{4.16} & \meanstd{0.70}{0.43} & \meanstd{0.67}{0.47} & \bestmeanstd{0.45}{0.28} \\
          & & ratio (\%) & 75.94 & 89.18 & 35.71 & 32.84 & - \\ \cmidrule{2-8}
          & \multirow{2}{*}{MAPE}
            & average (± std) & \meanstd{22.71}{4.91} & \meanstd{30.18}{9.28} & \meanstd{13.18}{4.12} & \meanstd{13.05}{4.71} & \bestmeanstd{10.71}{2.84} \\
          & & ratio (\%) & 52.84 & 64.51 & 18.74 & 17.93 & - \\
          \midrule
    
          \multirow{4}{*}{20 samples} & \multirow{2}{*}{MSE}
            & average (± std) & \meanstd{1.39}{0.60} & \meanstd{1.92}{1.50} & \meanstd{0.53}{0.16} & \meanstd{0.36}{0.18} & \bestmeanstd{0.30}{0.13} \\
          & & ratio (\%) & 78.42 & 84.38 & 43.40 & 16.67 & - \\ \cmidrule{2-8}
          & \multirow{2}{*}{MAPE}
            & average (± std) & \meanstd{18.88}{3.95} & \meanstd{20.11}{7.17} & \meanstd{11.87}{2.17} & \meanstd{9.50}{2.37} & \bestmeanstd{8.85}{2.23} \\
          & & ratio (\%) & 53.13 & 55.99 & 25.44 & 6.84 & - \\
          \midrule
    
          \multirow{4}{*}{40 samples} & \multirow{2}{*}{MSE}
            & average (± std) & \meanstd{0.52}{0.17} & \meanstd{0.60}{0.18} & \meanstd{0.49}{0.10} & \bestmeanstd{0.18}{0.06} & \meanstd{0.20}{0.06} \\
          & & ratio (\%) & 61.54 & 66.67 & 59.18 & -11.11 & - \\ \cmidrule{2-8}
          & \multirow{2}{*}{MAPE}
            & average (± std) & \meanstd{11.47}{1.90} & \meanstd{12.24}{1.78} & \meanstd{11.35}{1.16} & \bestmeanstd{6.52}{1.32} & \meanstd{6.85}{0.97} \\
          & & ratio (\%) & 40.28 & 44.04 & 39.65 & -5.06 & - \\
          \midrule
    
          \multirow{4}{*}{80 samples} & \multirow{2}{*}{MSE}
            & average (± std) & \meanstd{0.21}{0.08} & \meanstd{0.43}{0.08} & \meanstd{0.44}{0.08} & \bestmeanstd{0.09}{0.02} & \bestmeanstd{0.09}{0.03} \\
          & & ratio (\%) & 57.14 & 79.07 & 79.55 & 0.00 & - \\ \cmidrule{2-8}
          & \multirow{2}{*}{MAPE}
            & average (± std) & \meanstd{7.35}{1.24} & \meanstd{10.45}{0.83} & \meanstd{11.00}{0.90} & \meanstd{4.77}{0.52} & \bestmeanstd{4.63}{0.60} \\
          & & ratio (\%) & 37.01 & 55.69 & 57.91 & 2.94 & - \\
          \midrule
    
          \multirow{4}{*}{100 samples} & \multirow{2}{*}{MSE}
            & average (± std) & \meanstd{0.19}{0.05} & \meanstd{0.41}{0.06} & \meanstd{0.41}{0.05} & \meanstd{0.08}{0.01} & \bestmeanstd{0.07}{0.01} \\
          & & ratio (\%) & 63.16 & 82.93 & 82.93 & 12.50 & - \\ \cmidrule{2-8}
          & \multirow{2}{*}{MAPE}
            & average (± std) & \meanstd{7.01}{0.82} & \meanstd{10.61}{0.91} & \meanstd{10.72}{0.69} & \meanstd{4.26}{0.34} & \bestmeanstd{3.98}{0.36} \\
          & & ratio (\%) & 43.22 & 62.49 & 62.87 & 6.57 & - \\
          \bottomrule
        \end{tabular}
        \begin{tablenotes}
          \item[a] ``ratio'' indicates the relative improvement of CFKD-AFN over the competing method.
        \end{tablenotes}
      \end{threeparttable}
    }
  \label{tab: Fine-grained Performance Comparison for T3}
\end{table*}

\begin{table*}[tbp]
  \centering
  \caption{Fine-grained Performance Comparisons among Different Methods for Treatment 4}
  \resizebox{0.8\linewidth}{!}{
      \begin{threeparttable}
        \begin{tabular}{
          >{\centering\arraybackslash}m{2cm}
          >{\centering\arraybackslash}m{1.5cm}
          >{\centering\arraybackslash}m{2.2cm}
          >{\centering\arraybackslash}m{1.5cm}
          >{\centering\arraybackslash}m{1.5cm}
          >{\centering\arraybackslash}m{1.5cm}
          >{\centering\arraybackslash}m{1.5cm}
          >{\centering\arraybackslash}m{1.5cm}
        }
          \toprule
          & & & \textbf{HF} & \textbf{PFT} & \textbf{MF-TLNN} & \textbf{MF-DF} & \textbf{CFKD-AFN} \\
          \midrule
    
          \multirow{4}{*}{10 samples} & \multirow{2}{*}{MSE}
            & average (± std) & \meanstd{2.11}{1.24} & \meanstd{0.65}{0.32} & \meanstd{0.57}{0.32} & \meanstd{0.70}{0.49} & \bestmeanstd{0.47}{0.21} \\
          & & ratio (\%) & 77.73 & 27.69 & 17.54 & 32.86 & - \\ \cmidrule{2-8}
          & \multirow{2}{*}{MAPE}
            & average (± std) & \meanstd{22.93}{6.64} & \meanstd{12.80}{2.97} & \meanstd{11.83}{2.95} & \meanstd{12.86}{3.97} & \bestmeanstd{11.00}{2.03} \\
          & & ratio (\%) & 52.03 & 14.06 & 7.02 & 14.46 & - \\
          \midrule
    
          \multirow{4}{*}{20 samples} & \multirow{2}{*}{MSE}
            & average (± std) & \meanstd{1.40}{0.59} & \meanstd{0.56}{0.25} & \meanstd{0.47}{0.18} & \meanstd{0.46}{0.20} & \bestmeanstd{0.44}{0.18} \\
          & & ratio (\%) & 68.57 & 21.43 & 6.38 & 4.35 & - \\ \cmidrule{2-8}
          & \multirow{2}{*}{MAPE}
            & average (± std) & \meanstd{17.67}{4.25} & \meanstd{11.36}{2.53} & \meanstd{10.61}{1.94} & \meanstd{10.31}{1.96} & \bestmeanstd{10.10}{1.91} \\
          & & ratio (\%) & 42.84 & 11.09 & 4.81 & 2.04 & - \\
          \midrule
    
          \multirow{4}{*}{40 samples} & \multirow{2}{*}{MSE}
            & average (± std) & \meanstd{0.62}{0.27} & \meanstd{0.42}{0.13} & \meanstd{0.42}{0.14} & \meanstd{0.19}{0.08} & \bestmeanstd{0.18}{0.06} \\
          & & ratio (\%) & 70.97 & 57.14 & 57.14 & 5.26 & - \\ \cmidrule{2-8}
          & \multirow{2}{*}{MAPE}
            & average (± std) & \meanstd{11.74}{2.16} & \meanstd{10.05}{1.50} & \meanstd{10.06}{1.58} & \bestmeanstd{6.50}{1.32} & \meanstd{6.55}{0.95} \\
          & & ratio (\%) & 44.21 & 34.83 & 34.89 & -0.77 & - \\
          \midrule
    
          \multirow{4}{*}{80 samples} & \multirow{2}{*}{MSE}
            & average (± std) & \meanstd{0.22}{0.05} & \meanstd{0.38}{0.07} & \meanstd{0.39}{0.06} & \meanstd{0.09}{0.02} & \bestmeanstd{0.07}{0.01} \\
          & & ratio (\%) & 68.18 & 81.58 & 82.05 & 22.22 & - \\ \cmidrule{2-8}
          & \multirow{2}{*}{MAPE}
            & average (± std) & \meanstd{7.15}{0.71} & \meanstd{9.59}{0.92} & \meanstd{9.71}{0.77} & \meanstd{4.46}{0.50} & \bestmeanstd{4.01}{0.40} \\
          & & ratio (\%) & 43.92 & 58.19 & 58.70 & 10.09 & - \\
          \midrule
    
          \multirow{4}{*}{100 samples} & \multirow{2}{*}{MSE}
            & average (± std) & \meanstd{0.18}{0.05} & \meanstd{0.38}{0.06} & \meanstd{0.38}{0.06} & \meanstd{0.09}{0.02} & \bestmeanstd{0.07}{0.02} \\
          & & ratio (\%) & 61.11 & 81.58 & 81.58 & 22.22 & - \\ \cmidrule{2-8}
          & \multirow{2}{*}{MAPE}
            & average (± std) & \meanstd{6.60}{0.76} & \meanstd{9.56}{0.80} & \meanstd{9.74}{0.72} & \meanstd{4.35}{0.37} & \bestmeanstd{3.90}{0.38} \\
          & & ratio (\%) & 40.91 & 59.21 & 59.96 & 10.34 & - \\
          \bottomrule
        \end{tabular}
        \begin{tablenotes}
          \item[a] ``ratio'' indicates the relative improvement of CFKD-AFN over the competing method.
        \end{tablenotes}
      \end{threeparttable}
    }
  \label{tab: Fine-grained Performance Comparison for T4}
\end{table*}

\subsubsection{Fine-grained Performance Comparison} To further assess fine-grained performance of the methods, each method is evaluated with total high-fidelity sample sizes of 10, 20, 40, 80, and 100. We record the results averaged over 20 repetitions per setting to examine robustness across different sample sizes.
Tables~\ref{tab: Fine-grained Performance Comparison for T1}-\ref{tab: Fine-grained Performance Comparison for T4} present the fine-grained performance comparison results in terms of average accuracy and standard deviation (``std'').
Two main observations can be made based on these results.

\textbf{Observation 1: In the extremely small-sample setting, CFKD-AFN outperforms or performs comparably to existing methods.}
In the settings with 10 samples, CFKD-AFN performs better than HF, PFT, and MF-DF with improvements of 23.53\%-89.18\% in MSE and 11.51\%-64.51\% in MAPE, and its performance is comparable to that of MF-TLNN.
Moreover, this scenario further validates the earlier analysis that when the number of high-fidelity samples is very limited, the one-dimensional output in MF-DF fails to provide sufficient transfer information, which may result in inferior performance compared to MF-TLNN.

\textbf{Observation 2: CFKD-AFN exhibits robustness across varying scales of high-fidelity data.} Under limited data conditions (20, 40, 80, and 100 samples), CFKD-AFN consistently outperforms HF, PFT, and MF-TLNN, and generally outperforms or performs comparably to MF-DF.
This advantage stems from its multi-level knowledge distillation and adaptive fusion mechanism, which integrates the output prior, high-dimensional representative knowledge, and raw input information. By leveraging an adaptive attention mechanism for multi-level feature fusion, CFKD-AFN delivers predictions that are both accurate and robust.

\begin{table*}[tbp]
	\centering
	\caption{Results of the Ablation Study for Four Treatments}
    \resizebox{0.9\linewidth}{!}{
    	\begin{threeparttable}
    	\begin{tabular}
    		{
    			>{\centering\arraybackslash}m{1.5cm}
    			>{\centering\arraybackslash}m{1.5cm} 
    			>{\centering\arraybackslash}m{1.5cm} 
    			>{\centering\arraybackslash}m{2.5cm} 
    			>{\centering\arraybackslash}m{3.2cm} 
    			>{\centering\arraybackslash}m{2.5cm} 
    			>{\centering\arraybackslash}m{2.5cm}  
    		}
    		\toprule
    		&      &            & \textbf{CFKD-AFN w/o $\bm{y_{lh}}$} & \textbf{CFKD-AFN w/o $\bm{{\textbf{y}}_{feature}}$} & \textbf{CFKD-AFN w/o $\bm{\textbf{x}_h}$}          & \textbf{CFKD-AFN}       \\ 
    		\midrule
    		\multirow{4}{*}{Treatment 1} & \multirow{2}{*}{MSE}  & average       & 0.31                 & 0.31                      & 0.33                       & \textbf{0.30}  \\
    		&      & ratio (\%) & 3.23                 & 3.23                      & 9.09                       & -             \\ \cmidrule{2-7} 
    		& \multirow{2}{*}{MAPE} & average       & 8.18                 & \textbf{7.93}             & 8.30 & 8.06          \\
    		&      & ratio (\%) & 1.47                 & -1.64                     & 2.89                       & -             \\ \midrule
    		\multirow{4}{*}{Treatment 2} & \multirow{2}{*}{MSE}  & average       & 0.22                 & 0.20                       & 0.23                       & \textbf{0.18} \\
    		&      & ratio (\%) & 18.18                & 10.00                        & 21.74                      & -             \\ \cmidrule{2-7} 
    		& \multirow{2}{*}{MAPE} & average       & 6.88                 & 6.39                      & 6.60                        & \textbf{6.18} \\
    		&      & ratio (\%) & 10.17                & 3.29                      & 6.36                       & -             \\ \midrule
    		\multirow{4}{*}{Treatment 3} & \multirow{2}{*}{MSE}  & average       & 0.24                 & 0.29                      & \textbf{0.23}              & 0.24          \\
    		&      & ratio (\%) & 0.00                    & 17.24                     & -4.35                      & -             \\ \cmidrule{2-7} 
    		& \multirow{2}{*}{MAPE} & average       & 7.23                 & 7.14                      & 6.98                       & \textbf{6.82} \\
    		&      & ratio (\%) & 5.67                 & 4.48                      & 2.29                       & -             \\ \midrule
    		\multirow{4}{*}{Treatment 4} & \multirow{2}{*}{MSE}  & average       & 0.29                 & 0.20                       & \textbf{0.18}                       & \textbf{0.18} \\
    		&      & ratio (\%) & 37.93                & 10.00                        & 0.00                          & -             \\ \cmidrule{2-7} 
    		& \multirow{2}{*}{MAPE} & average       & 7.51                 & 6.46                      & 6.18                       & \textbf{6.06} \\
    		&      & ratio (\%) & 19.31                & 6.19                      & 1.94                       & -             \\ 
    		\bottomrule
    	\end{tabular}
    	\begin{tablenotes}
    		\item[a]  ``ratio'' indicates the relative improvement of CFKD-AFN over the competing method.
    	\end{tablenotes}
    	\end{threeparttable}
    }
	\label{tab: ablation experiments}
\end{table*}

\subsection{Ablation Study}
To assess the contribution of multi-level information in CFKD-AFN to predictive accuracy, an ablation study is conducted with three variant models: CFKD-AFN w/o $y_{lh}$ (retaining only the ${\textbf{y}}_{feature}$ and $\textbf{x}_h$), CFKD-AFN w/o ${\textbf{y}}_{feature}$ (retaining only the $y_{lh}$ and $\textbf{x}_h$), and CFKD-AFN w/o $\textbf{x}_h$ (retaining only the $y_{lh}$ and ${\textbf{y}}_{feature}$).
Apart from the transferred information, all other configurations of the variant models are identical to CFKD-AFN.
For evaluation, 50 datasets are sampled from the second high-fidelity pool, with each dataset size drawn uniformly from [10,100], and the average predictive performance of the four models is compared.
Table~\ref{tab: ablation experiments} presents the results of the ablation experiments.
Based on these results, two observations can be made.

\textbf{Observation 1: The three types of predictive information make critical contributions to model performance.}
The results show that the full CFKD-AFN model generally performs at or near the top.
Specifically, in the prediction tasks for Treatments 2 and 4, CFKD-AFN outperforms other baselines, with only slight performance drops compared to CFKD-AFN w/o ${\textbf{y}}_{feature}$ in Treatment 1 and CFKD-AFN w/o $\textbf{x}_h$ in Treatment 3. 

\textbf{Observation 2: The importance of each information source varies across tasks.}
Among the variant models, for example, CFKD-AFN w/o $y_{lh}$ performs best in Treatment 1, while CFKD-AFN w/o ${\textbf{y}}_{feature}$ performs best in both Treatments 1 and 2; CFKD-AFN w/o $\textbf{x}_h$ exhibits the top performance in Treatments 3 and 4.
These results reveal that removing any one of the information sources may degrade the model's generalization ability in specific scenarios.
Therefore, integrating multi-level information in the full CFKD-AFN model provides a more robust solution across all tasks.

\section{Exploration of Model Interpretability}
\label{Exploration of Model Interpretability}
In personalized medicine, it is crucial to ensure that the decision-making process of machine learning models is transparent to support physicians in practice.
To this end, an interpretable variant (iCFKD-AFN) is extended from CFKD-AFN that employs disentangled representation learning to identify {\color{black}partially separated feature-attribution} patterns associated with treatment outcomes. 
To gain deeper insights, two latent vectors in iCFKD-AFN are visualized to explore their {\color{black}feature-attribution patterns} and compare the predictive accuracy of iCFKD-AFN with CFKD-AFN.

\begin{figure*}[!t]
	\centering
	\includegraphics[width=0.7\textwidth]{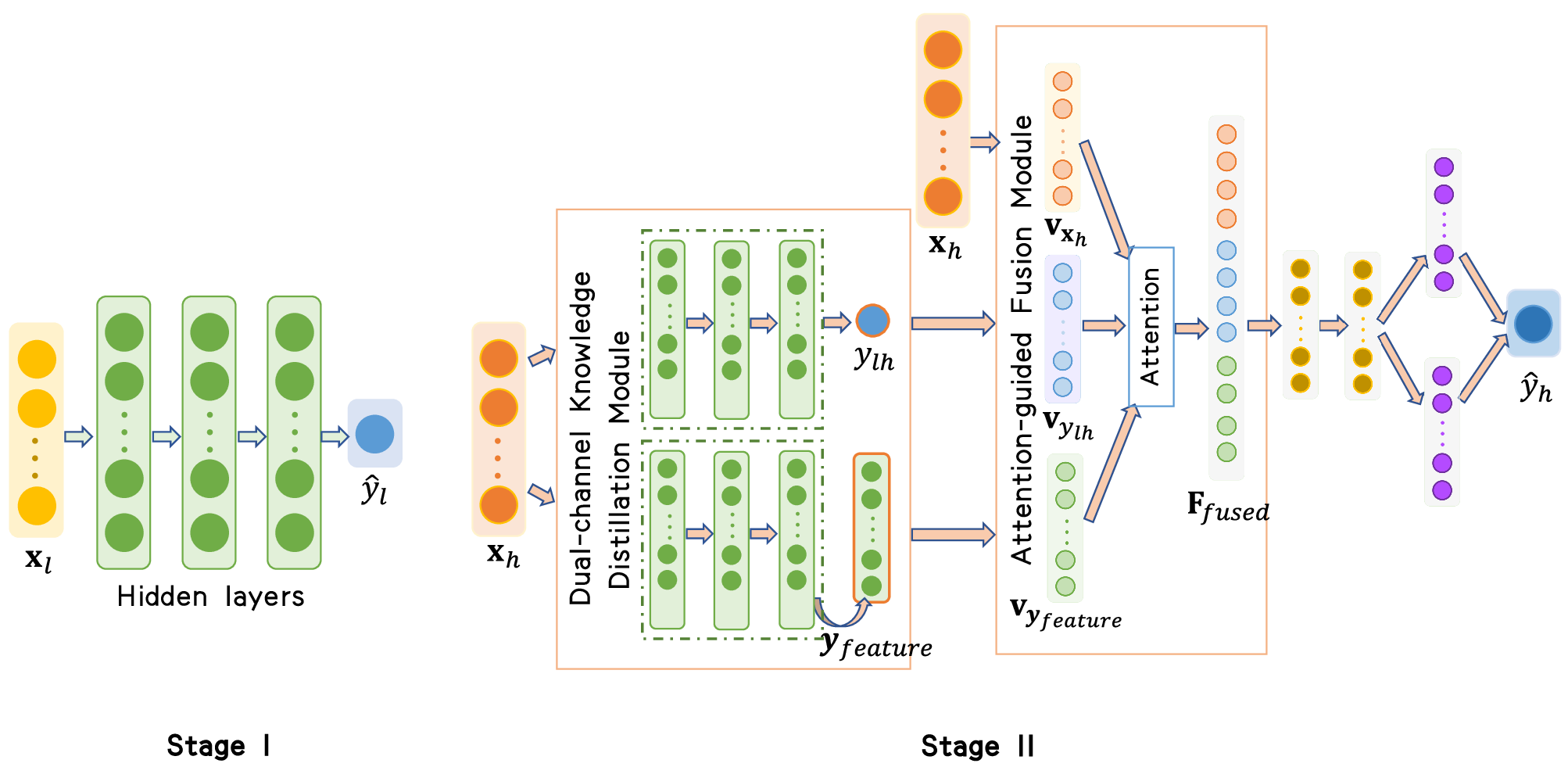} 
	\caption{The architecture of iCFKD-AFN.}
	\label{fig: iCFKD-AFN} 
\end{figure*}

\subsection{The iCFKD-AFN Variant}
Based on the final hidden layer representations of CFKD-AFN, two additional disentangled vectors are introduced that are linked to the last hidden layer and regularized using a mutual information constraint to achieve effective factor separation, as illustrated in Fig.~\ref{fig: iCFKD-AFN}. The last hidden layer is chosen because it provides the most task-relevant and high-level representations while also keeping the model simple. {\color{black}Let \(\mathbf{h}_i\) denote the last hidden representation of the \(i\)-th sample. The two vectors and the final prediction are defined as
\[\text{vec}_{k,i}=\phi(\mathbf{w}_k\mathbf{h}_i+\mathbf{b}_k),\ k\in\{1,2\},
\]
\[	\hat y_{h,i}=\mathbf{w}_o\operatorname{Concat}
	(\text{vec}_{1,i}, \text{vec}_{2,i})+\mathbf{b}_o,\]
where \(\mathbf{w}_k\) and \(\mathbf{b}_k\) are the projection parameters, \(\phi(\cdot)\) is the activation function, and \(\mathbf{w}_o\) and \(\mathbf{b}_o\) are the parameters of the final layer.}

The model is then trained with the loss function in Eq.~(\ref{loss function}).
This regularization encourages separation between the two latent vectors, allowing them to capture {\color{black}different feature patterns.}  
\begin{equation}
	\label{loss function}
	Loss = \frac{1}{t_h}\cdot \sum_{i=1}^{t_h} \left( y_{h,i} - \hat{y}_{h,i} \right)^2 + \beta\cdot MI(\text{vec}_1, \text{vec}_2),
\end{equation}
where $MI(\cdot)$ represents the mutual information function, and $\beta$ is the weight parameter to balance prediction accuracy and model interpretability. 

The iCFKD-AFN method introduces two additional key hyperparameters compared with CFKD-AFN: the weight parameter $\beta$ and the dimensionality of the decoupled vector $d$.
These two hyperparameters are tuned using the grid search with $\beta\in [100,500,1000]$ and $d\in [8,16,32]$, while keeping other hyperparameters identical to those in CFKD-AFN.
Taking Treatment 1 as an example, the optimal hyperparameters are determined as $\beta=500$ and $d=32$.

\subsection{Experiments on Model Interpretability}
\begin{figure*}[tbp]
	\centering
	\subfloat[\scriptsize 50 high-fidelity samples]{\includegraphics[width=0.495\textwidth]{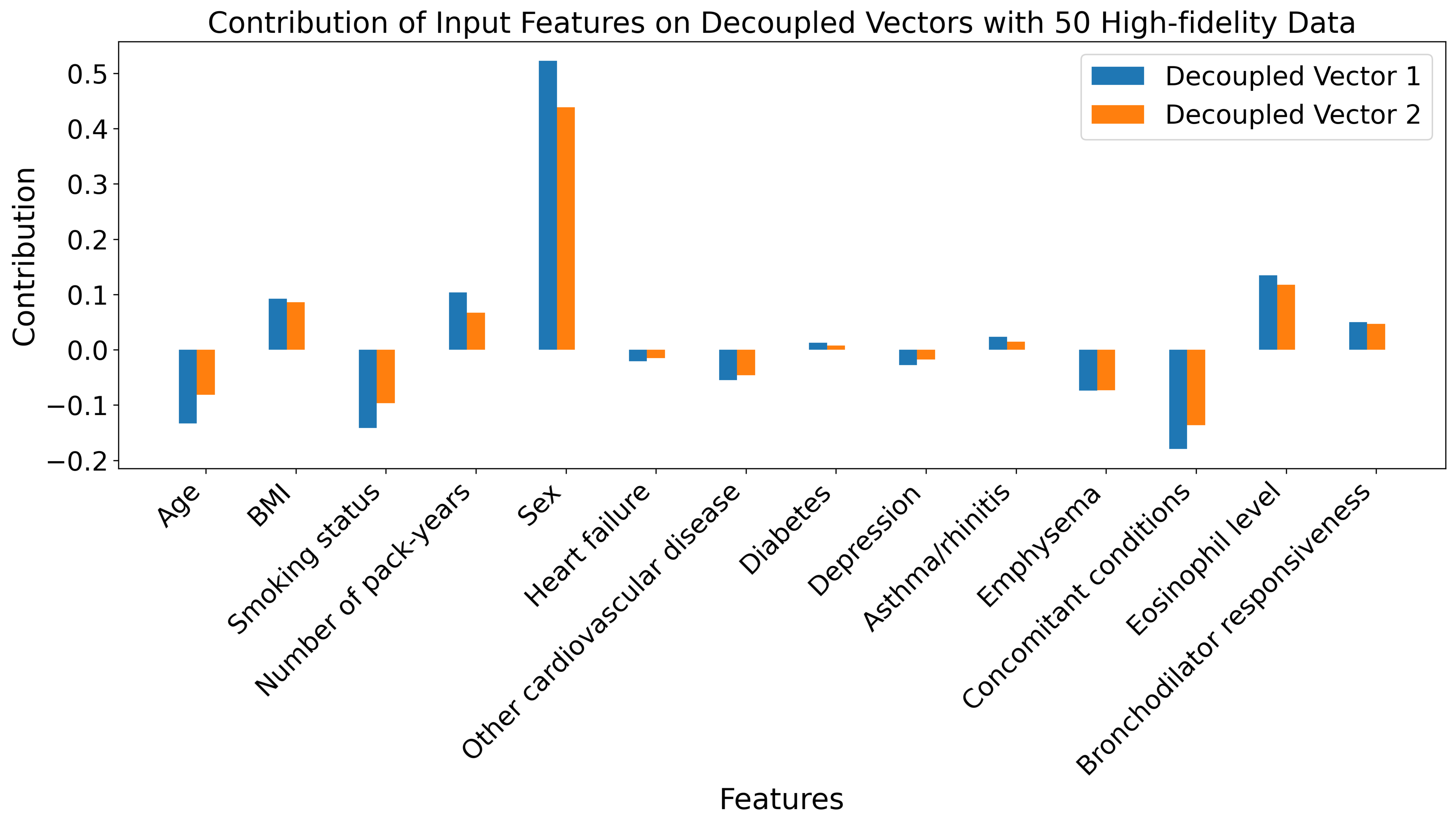}%
		\label{fig: 50 data}}
	\subfloat[\scriptsize 100 high-fidelity samples]{\includegraphics[width=0.495\textwidth]{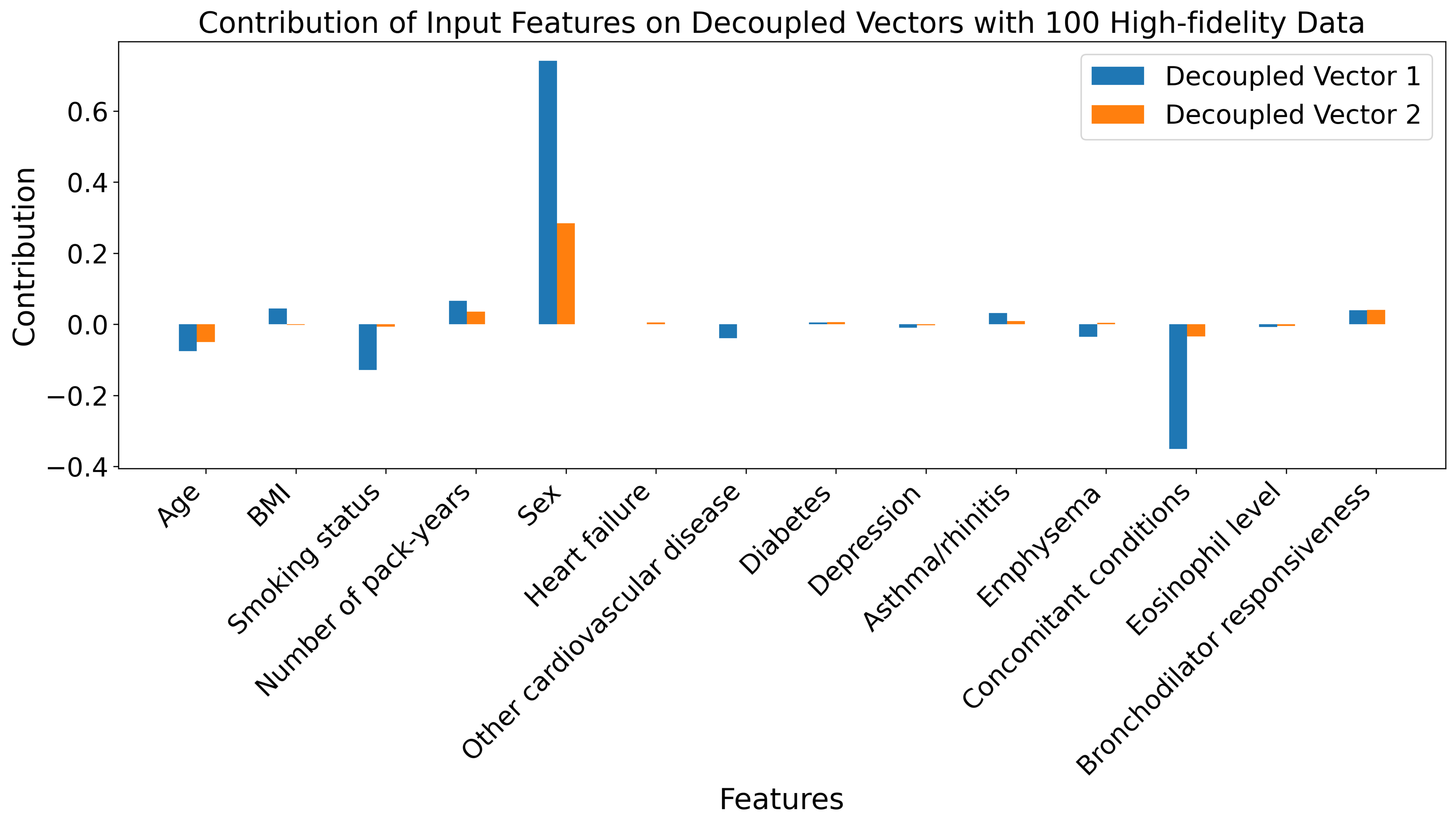}%
		\label{fig: 100 data}}
	\hfil
	\subfloat[\scriptsize 200 high-fidelity samples]{\includegraphics[width=0.495\textwidth]{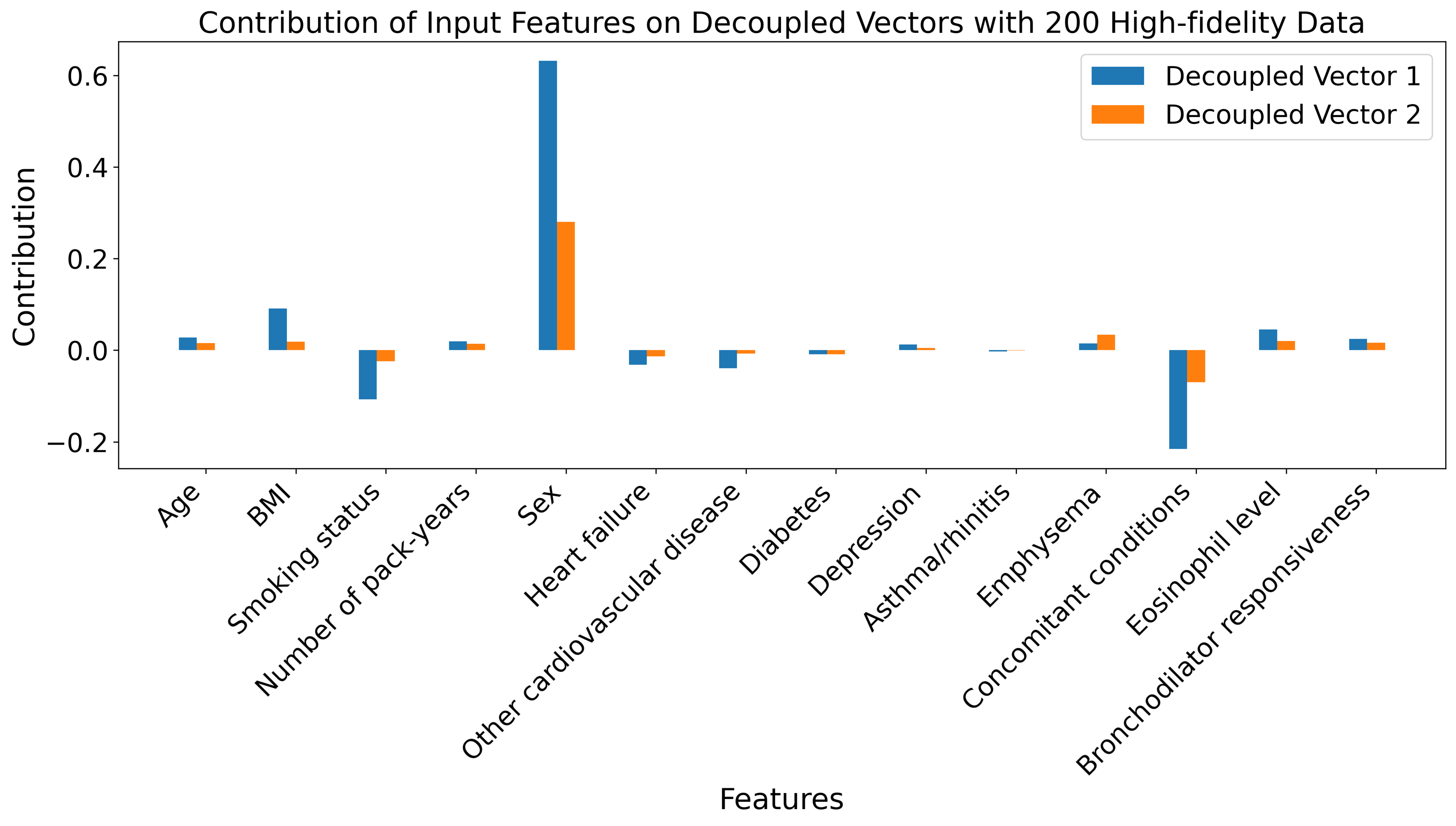}%
		\label{fig: 200 data}}
	\subfloat[\scriptsize 400 high-fidelity samples]{\includegraphics[width=0.495\textwidth]{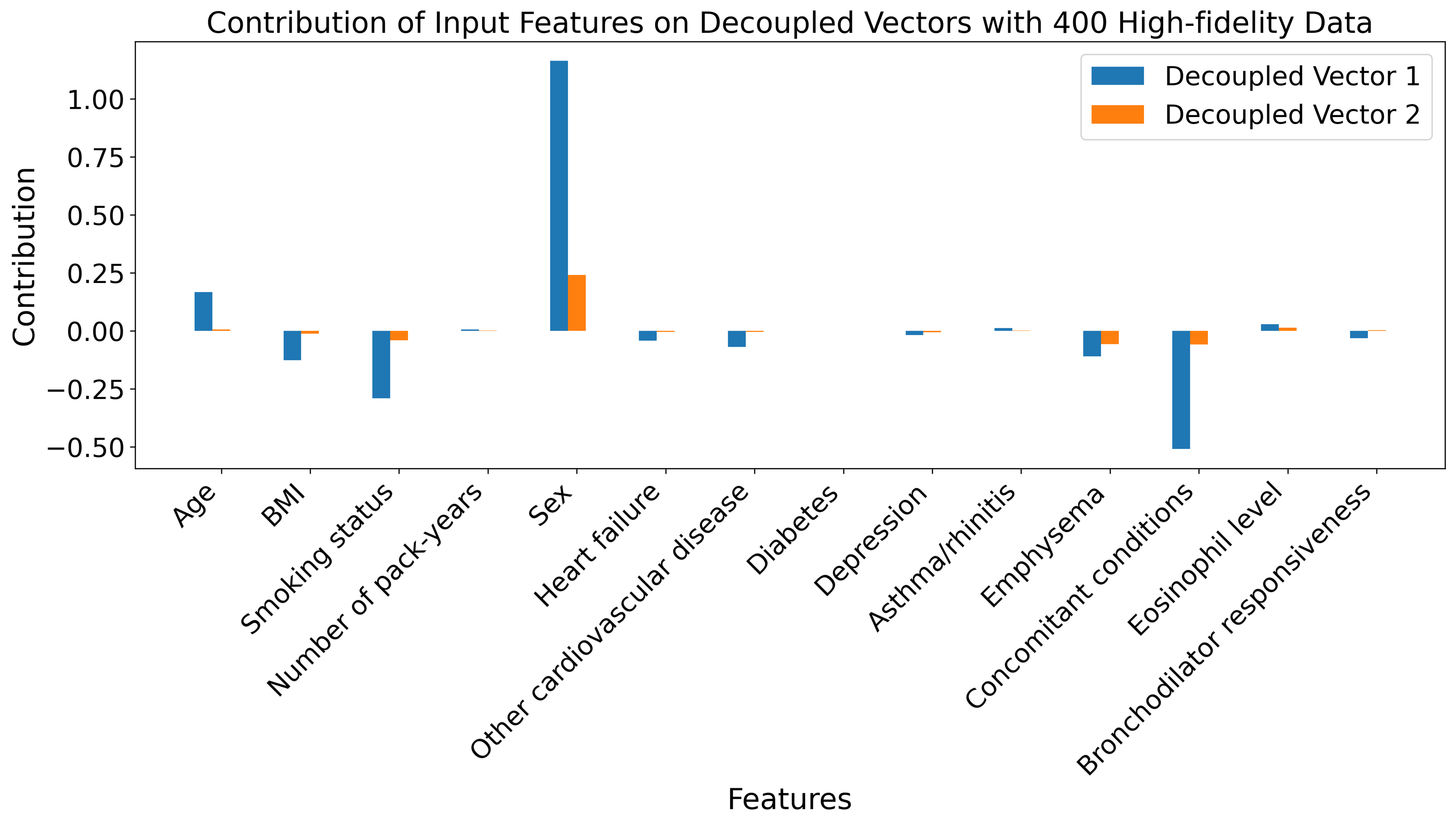}%
		\label{fig: 400 data}}
	\caption{Contributions of the input features on the decoupled vectors with different amounts of high-fidelity data.}
	\label{fig: iCFKD-AFN experiment 1}
\end{figure*}

Our preliminary experiments reveal that under limited high-fidelity data, iCFKD-AFN exhibits insufficient statistical power for identification of outcome-associated feature patterns, as the learned latent representations demonstrate high variance and unstable factor loadings across different sample sizes.
Therefore, we generate additional 5000 high-fidelity samples and evaluate the model’s interpretability under relatively large data scales (e.g., 200 samples).
First, the mutual information between the two connected vectors is estimated using the Donsker-Varadhan method \cite{belghazi2018mutual}.
Simultaneously, the Gradient×Input technique~\cite{nielsen2022robust} is employed to compute the contribution weights of input features to each vector.
Finally, the latent {\color{black}representation} distribution of the vectors is analyzed through visualization as shown in Fig.~\ref{fig: iCFKD-AFN experiment 1}.

\textbf{Observation 1: The interpretability of iCFKD-AFN improves as the amount of high-fidelity data increases.} With only 50 samples, the feature contributions of the two latent vectors fluctuate drastically. The “core feature sets” of the two vectors have not yet formed, resulting in unstable and ambiguous latent meanings. In contrast, with 100, 200, and 400 samples, the contribution patterns gradually stabilize in both sign and relative magnitude. For example, the association of concomitant conditions with vector 1 and the association of sex with vector 2 emerge, and the differences between the two vectors become clearer.

\textbf{Observation 2: When the sample size is at least 100, the two vectors show partial decoupling with distinct focuses on the relative weighting and scope of feature associations.}
In Fig.~\ref{fig: iCFKD-AFN experiment 1},
sex shows the largest attribution for both vectors, indicating that it is important in both latent representations. However, vector 1 exhibits a broader attribution pattern involving concomitant conditions and several disease-related features, whereas vector 2 is more strongly concentrated on sex, with relatively small attributions from the remaining features. 
The partial separation suggests that the two vectors emphasize different groups of correlated clinical features, where demographic variables influence both baseline health status and treatment outcomes.

\subsection{Comparison of CFKD-AFN and iCFKD-AFN in terms of Accuracy}
A comparison of the prediction accuracy of CFKD-AFN and iCFKD-AFN is also presented, using 50, 100, 200, and 400 high-fidelity samples.
For each experiment, 20 repetitions are conducted to calculate the average MSE and MAPE.
Table~\ref{tab: comparison of CFKD-AFN and iCFKD-AFN} shows the comparison results. 

\begin{table}[]
	\centering
	\caption{Prediction Accuracy Comparisons between CFKD-AFN and iCFKD-AFN with Different Amounts of High-fidelity Data}
	\begin{tabular}
		{
			p{1.3cm}
			>{\centering\arraybackslash}m{0.45cm} 
			>{\centering\arraybackslash}m{0.5cm} 
			>{\centering\arraybackslash}m{0.45cm} 
			>{\centering\arraybackslash}m{0.5cm}
			>{\centering\arraybackslash}m{0.45cm} 
			>{\centering\arraybackslash}m{0.5cm} 
			>{\centering\arraybackslash}m{0.45cm} 
			>{\centering\arraybackslash}m{0.5cm}
		}
		\toprule
		& \multicolumn{2}{c}{\textbf{50}} & \multicolumn{2}{c}{\textbf{100}} & \multicolumn{2}{c}{\textbf{200}} & \multicolumn{2}{c}{\textbf{400}} \\ 
		\midrule
		& {MSE} & {MAPE} & {MSE} & {MAPE} & {MSE} & {MAPE} & {MSE} & {MAPE}  \\ 
		\midrule
		CFKD-AFN & 0.31 & 8.00 & 0.10 & 4.78 & 0.06 & 3.64 & 0.04 & 2.98 \\
		\midrule
		iCFKD-AFN & 0.77 & 14.03 & 0.65 & 12.67 & 0.53 & 11.07 & 0.31 & 7.87 \\
		\bottomrule
	\end{tabular}
	\label{tab: comparison of CFKD-AFN and iCFKD-AFN}
\end{table}

\textbf{Observation 1: Under the same data scale, iCFKD-AFN exhibits lower accuracy than CFKD-AFN.} This is because the training objective of iCFKD-AFN combines the mean squared error with a decoupling regularization term. This design enhances model interpretability at the expense of some predictive accuracy. Notably, as the sample size increases, the accuracy of iCFKD-AFN steadily improves, suggesting that the model benefits from larger datasets while maintaining its interpretability. In summary, CFKD-AFN prioritizes predictive accuracy for trial-based treatment outcome prediction, whereas iCFKD-AFN supports exploratory feature-pattern analysis at the cost of some predictive performance.

\section{Conclusion}
\label{Conclusion}
This work focuses on a critical challenge in personalized medicine, i.e., treatment outcome prediction for small-sample and rare patient groups based on costly trial data.
The cross-fidelity knowledge distillation and adaptive fusion network is proposed to effectively transfer knowledge from abundant low-fidelity simulation data to limited high-fidelity trial data.
By integrating a dual-channel knowledge distillation module with an attention-guided fusion mechanism,  CFKD-AFN {\color{black}provides treatment-specific outcome predictions and} achieves lower MSE and MAPE than the compared methods in most evaluated settings for the COPD case study.
Moreover, the interpretable extension iCFKD-AFN provides {\color{black}exploratory analysis of feature-attribution patterns.}

Several promising directions remain for future research.
First, more effective cross-fidelity transfer strategies are needed to improve both predictive accuracy and interpretability when high-fidelity data are limited.
Second, techniques, such as contrastive learning~\cite{WuFCL0T22}, may help align low- and high-fidelity representations and reduce the effects of distribution discrepancies.
Other strategies, such as active learning~\cite{HeLHT23}, are also worth exploring to guide the efficient selection of high-fidelity samples.
Additionally, large language model- and optimization-based condensation may generate synthetic data to supplement scarce trial data~\cite{Wu2025WWW,Wu2024TKDE}.
Finally, validation using real-world trial data is needed to assess the practical applicability of our methods.

\begin{appendices}

\begin{table*}[]
	\centering
	\caption{Distributions for High- and Low-Fidelity Data Generation}
    \resizebox{0.8\linewidth}{!}{
    \begin{threeparttable}
    	\begin{tabular}{cll}
    		\toprule
    		\textbf{Feature}                         & \multicolumn{1}{c}{\textbf{High-fidelity data}}                      & \multicolumn{1}{c}{\textbf{Low-fidelity data}}      \\ 
    		\midrule
    		\textbf{Age}                             & $\text{Round}\left(\text{Normal}(64,\,7^2)\,\big|_{[40,85]}\right)$  & $\text{Uniform}\{40,41,...,85\}$                             \\
    		\textbf{BMI}                             & $\text{Categorical}(p_1=0.15,\, p_2=0.61,\, p_3=0.24)$               & $\text{Categorical}(p_1=1/3,\, p_2=1/3,\, p_3=1/3)$ \\
    		\textbf{Smoking status}                  & $\text{Bernoulli}(p=0.39)$                                           & $\text{Bernoulli}(p=0.5)$                           \\
    		\textbf{Number    of  pack-years}        & $\text{Normal}(44,\,15^2)\,\big|_{[10,89]}$ & $\text{Uniform}\{10,11,...,89\}$                             \\
    		\textbf{Sex (Female)}                    & $\text{Bernoulli}(p=0.27)$                                           & $\text{Bernoulli}(p=0.5)$                           \\
    		\textbf{Heart    failure}                & $\text{Bernoulli}(p=0.05)$                                           & $\text{Bernoulli}(p=0.5)$                           \\
    		\textbf{Other    cardiovascular disease} & $\text{Bernoulli}(p=0.12)$                                           & $\text{Bernoulli}(p=0.5)$                           \\
    		\textbf{Diabetes}                        & $\text{Bernoulli}(p=0.11)$                                           & $\text{Bernoulli}(p=0.5)$                           \\
    		\textbf{Depression}                      & $\text{Bernoulli}(p=0.085)$                                          & $\text{Bernoulli}(p=0.5)$                           \\
    		\textbf{Asthma/rhinitis}                 & $\text{Bernoulli}(p=0.047)$                                          & $\text{Bernoulli}(p=0.5)$                           \\
    		\textbf{Emphysema}                       & $\text{Bernoulli}(p=0.53)$                                           & $\text{Bernoulli}(p=0.5)$                           \\
    		\textbf{Concomitant conditions}          & $\text{Bernoulli}(p=0.56)$                                           & $\text{Bernoulli}(p=0.5)$                           \\
    		\textbf{Eosinophil level}                & $\text{Bernoulli}(p=0.22)$                                           & $\text{Bernoulli}(p=0.5)$                           \\
    		\textbf{Bronchodilator responsiveness}   & $\text{Normal}(23,\,5^2)\,\big|_{[8,38]}$  & $\text{Uniform}\{8,9,...,38\}$                             \\ 
    		\bottomrule
    	\end{tabular}
    		\begin{tablenotes}
    		\item[a]  The subscripted intervals indicate the truncation ranges of the corresponding normal distributions.
    	\end{tablenotes}
    	\end{threeparttable}
    }
	\label{tab: data generation}
\end{table*}

\begin{table*}[!h]
	\centering
	\caption{Hyperparameter Settings for Low-fidelity Models}
    \resizebox{0.8\linewidth}{!}{
    	\begin{tabular}{ccccc}
    		\toprule
    		\textbf{Hyperparameter}              & \textbf{Treatment 1}         & \textbf{Treatment 2}     & \textbf{Treatment 3}      & \textbf{Treatment 4}          \\ \midrule
    		\textbf{Dimensions of hidden layers} & {[}256,128,64,32,16,16,16{]} & {[}128,64,32,16,16,16{]} & {[}128, 64, 32, 16, 16{]} & {[}128, 64, 32, 16, 16, 16{]} \\
    		\textbf{Learning rate}               & 0.01                         & 0.01                     & 0.01                      & 0.01                          \\
    		\textbf{Batch size}                  & 32                           & 32                       & 16                        & 32                            \\
    		\textbf{Activation function}         & LeakyReLU                    & LeakyReLU                & ReLU                      & LeakyReLU                     \\ \bottomrule
    	\end{tabular}
    }
	\label{tab: hyperparameters for low-fidelity model}
\end{table*}

\begin{table*}[!h]
	\centering
	\caption{Hyperparameter Settings for High-fidelity Models}
    \resizebox{0.9\linewidth}{!}{
    	\begin{tabular}{cccccc}
    		\toprule
    		\textbf{Algorithm}                & \textbf{Hyperparameter}                                                                         & \textbf{Treatment 1}         & \textbf{Treatment 2}         & \textbf{Treatment 3}         & \textbf{Treatment 4}         \\ \midrule
    		\multirow{2}{*}{\textbf{HF}}      & Dimensions of hidden layers                                                                     & {[}32, 16, 16, 16, 16, 16{]} & {[}32, 16, 16, 16, 16, 16{]} & {[}32, 16, 16, 16, 16, 16{]} & {[}32, 16, 16, 16, 16, 16{]} \\
    		& Learning rate                                                                                   & 0.01                         & 0.01                         & 0.01                         & 0.01                         \\ \midrule
    		\multirow{2}{*}{\textbf{PFT}}     & Number of unfrozen layers                                                                       & 2                            & 2                            & 2                            & 1                            \\
    		& Learning rate                                                                                   & 0.01                         & 0.01                         & 0.01                         & 0.01                         \\ \midrule
    		\multirow{5}{*}{\textbf{MF-TLNN}} & \begin{tabular}[c]{@{}c@{}}Number of unfrozen layers \\ for the low-fidelity model\end{tabular} & 2                            & 1                            & 1                            & 1                            \\
    		& \begin{tabular}[c]{@{}c@{}}Number of unfrozen layers\\ for the autoencoder model\end{tabular}  & 2                            & 2                            & 1                            & 1                            \\
    		& Learning rate                                                                                   & 0.01                         & 0.01                         & 0.01                         & 0.01                         \\ \midrule
    		\multirow{2}{*}{\textbf{MF-DF}}   & Dimensions of hidden layers                                                                     & {[}128, 64, 32, 16{]}        & {[}128, 64{]}                & {[}256, 128{]}               & {[}256, 128{]}               \\
    		& Learning rate                                                                                   & 0.01                         & 0.01                         & 0.01                         & 0.01                         \\ \midrule
    		\multirow{4}{*}{\textbf{CFKD-AFN}} & Dimensions of hidden layers                                                                     & {[}32,16,16{]}               & {[}32, 16, 16, 16, 16{]}     & {[}64, 32, 16, 16, 16{]}     & {[}64, 32, 16, 16, 16, 16{]} \\
    		& \begin{tabular}[c]{@{}c@{}}Representation dimension \\ after linear transformation\end{tabular} & 32                           & 32                           & 16                           & 16                           \\
    		& Learning rate                                                                                   & 0.01                         & 0.01                         & 0.01                         & 0.01                         \\ \bottomrule
    	\end{tabular}
    }
	\label{tab:Hyperparameter Settings for high-fidelity Models}
\end{table*}

\section{Low-fidelity COPD Simulation Modeling}
\label{app:Low-fidelity COPD Simulation Modeling}
The high-fidelity COPD model is built using the discrete-event simulation. The simulation process is as follows. It begins by randomly selecting patients from the initial population and initializing system state variables according to the specified treatment regimen. Using regression equations based on patient and disease characteristics, the model predicts the timing of three types of events—acute exacerbation, pneumonia, or death—and selects the earliest event, updating intermediate outcome variables (lung function first, followed by other outcomes) to reflect the direct impact of the event on health status. Based on the updated outcomes, the model then predicts the next event and iterates until patient death or lung function falls below 0.2 liters. If the interval between events exceeds one year, retrospective simulation is performed for the intervening years to generate annual intermediate outcomes, and QALYs are calculated. Each patient is simulated 1,000 times, and the average QALYs are taken as the final results. 

We employ the following two strategies to obtain a low-fidelity simulation: (1) for event intervals exceeding one year, no retrospective simulation is performed; instead, the intermediate outcome values from the preceding event are used to fill in the missing years, reducing model complexity; (2) the number of simulations per patient is reduced to 100, and the average is used to calculate the low-fidelity QALYs.

\section{High- and Low-fidelity Data Generation}
\label{app:High- and Low-fidelity Data Generation}
To simulate differences between source and target domains in practice, high-fidelity data follow the recommendations in \cite{hoogendoorn2019broadening, du2024contextual}. We use the Uniform distribution, the Bernoulli distribution with parameter 0.5, and the Categorical distribution with equal class probabilities to generate low-fidelity samples. This setting reflects the absence of reliable prior knowledge about the corresponding real-world feature distributions. The data distributions are shown in Table \ref{tab: data generation}.

\section{Hyperparameter Settings}
\label{app: Hyperparameters Setting}
To obtain the optimal low-fidelity model for each treatment, the following grid search is conducted. We configure the network depth to range from 5 to 7 layers. The number of neurons in the first layer is selected from [512, 256, 128], with subsequent layers halved progressively until reaching a minimum of 16 neurons. The activation functions are chosen from [ReLU, LeakyReLU], the batch size from [16, 32, 64], and the learning rate from [0.01, 0.001, 0.0001]. The optimal hyperparameter settings for each treatment are shown in Table \ref{tab: hyperparameters for low-fidelity model}.

For the hyperparameters in high-fidelity models, all methods use ReLU activation, a batch size of 8, and a learning rate range of [0.01, 0.001] for fairness. Beyond these shared settings, tailored grid searches are applied: HF and MF-DF explore network depths of [2–6] with first-layer dimensions from [256, 128, 64, 32]; PFT sets the number of unfrozen layers to [1, 2]; MF-TLNN applies the same range to both the low-fidelity model and autoencoder; CFKD-AFN follows the MF-DF search space with an additional range of [16, 32] for the representation dimension after linear transformation. Table \ref{tab:Hyperparameter Settings for high-fidelity Models} summarizes the hyperparameter settings for high-fidelity models.

\end{appendices}

\bibliographystyle{IEEEtran} % IEEE格式样式
\bibliography{library}

@article{liu2024semi,
	title={Semi-supervised contrastive learning for time series classification in healthcare},
	author={Liu, Xiaofeng and Liu, Zhihong and Li, Jie and Zhang, Xiang},
	journal={IEEE Transactions on Emerging Topics in Computational Intelligence},
	volume={9},
	number={1},
	pages={318--331},
	year={2025},
	publisher={IEEE}
}

@article{alvi2022long,
	title={A long short-term memory based framework for early detection of mild cognitive impairment from {EEG} signals},
	author={Alvi, Ashik Mostafa and Siuly, Siuly and Wang, Hua},
	journal={IEEE Transactions on Emerging Topics in Computational Intelligence},
	volume={7},
	number={2},
	pages={375--388},
	year={2023},
	publisher={IEEE}
}

@article{yang2021robust,
	title={Robust collaborative learning of patch-level and image-level annotations for diabetic retinopathy grading from fundus image},
	author={Yang, Yehui and Shang, Fangxin and Wu, Binghong and Yang, Dalu and Wang, Lei and Xu, Yanwu and Zhang, Wensheng and Zhang, Tianzhu},
	journal={IEEE Transactions on Cybernetics},
	volume={52},
	number={11},
	pages={11407--11417},
	year={2022},
	publisher={IEEE}
}

@article{wang2023precision,
	title={Precision medicine: Disease subtyping and tailored treatment},
	author={Wang, Richard C and Wang, Zhixiang},
	journal={Cancers},
	volume={15},
	number={15},
	pages={3837},
	year={2023},
	publisher={MDPI}
}

@article{turki2019clinical,
	title={Clinical intelligence: New machine learning techniques for predicting clinical drug response},
	author={Turki, Turki and Wang, Jason TL},
	journal={Computers in Biology and Medicine},
	volume={107},
	pages={302--322},
	year={2019},
	publisher={Elsevier}
}

@article{zhang2020errors,
	title={Errors-in-variables modeling of personalized treatment-response trajectories},
	author={Zhang, Guangyi and Ashrafi, Reza A and Juuti, Anne and Pietil{\"a}inen, Kirsi and Marttinen, Pekka},
	journal={IEEE Journal of Biomedical and Health Informatics},
	volume={25},
	number={1},
	pages={201--208},
	year={2021},
	publisher={IEEE}
}

@article{kuenzi2020predicting,
	title={Predicting drug response and synergy using a deep learning model of human cancer cells},
	author={Kuenzi, Brent M and Park, Jisoo and Fong, Samson H and Sanchez, Kyle S and Lee, John and Kreisberg, Jason F and Ma, Jianzhu and Ideker, Trey},
	journal={Cancer Cell},
	volume={38},
	number={5},
	pages={672--684},
	year={2020},
	publisher={Elsevier}
}

@article{du2024contextual,
	title={A contextual ranking and selection method for personalized medicine},
	author={Du, Jianzhong and Gao, Siyang and Chen, Chun-Hung},
	journal={Manufacturing \& Service Operations Management},
	volume={26},
	number={1},
	pages={167--181},
	year={2024},
	publisher={INFORMS}
}

@article{ayer2016heterogeneity,
	title={Heterogeneity in women’s adherence and its role in optimal breast cancer screening policies},
	author={Ayer, Turgay and Alagoz, Oguzhan and Stout, Natasha K and Burnside, Elizabeth S},
	journal={Management Science},
	volume={62},
	number={5},
	pages={1339--1362},
	year={2016},
	publisher={INFORMS}
}

@article{ghane2025multi,
	title={Multi-fidelity data fusion for inelastic woven composites: Combining recurrent neural networks with transfer learning},
	author={Ghane, Ehsan and Fagerstr{\"o}m, Martin and Mirkhalaf, Mohsen},
	journal={Composites Science and Technology},
	volume={267},
	pages={111163},
	year={2025},
	publisher={Elsevier}
}

@article{yue2024improving,
	title={Improving {RSW} nugget diameter prediction method: Unleashing the power of multi-fidelity neural networks and transfer learning},
	author={Yue, Zhong-Jie and Chen, Qiu-Ren and Bao, Zu-Guo and Huang, Li and Tan, Guo-Bi and Hou, Ze-Hong and Li, Mu-Shi and Huang, Shi-Yao and Zhao, Hai-Long and Kong, Jing-Yu and others},
	journal={Advances in Manufacturing},
	volume={12},
	number={3},
	pages={409--427},
	year={2024},
	publisher={Springer}
}

@article{buterez2024transfer,
	title={Transfer learning with graph neural networks for improved molecular property prediction in the multi-fidelity setting},
	author={Buterez, David and Janet, Jon Paul and Kiddle, Steven J and Oglic, Dino and Li{\'o}, Pietro},
	journal={Nature Communications},
	volume={15},
	number={1},
	pages={1517},
	year={2024},
	publisher={Nature Publishing Group UK London}
}

@article{tang2025multi,
	title={Multi-fidelity modeling method based on adaptive transfer learning},
	author={Tang, Fazhi and Li, Yubo and Huang, Jun and Liu, Feng},
	journal={Information Fusion},
	volume={120},
	pages={103045},
	year={2025},
	publisher={Elsevier}
}

@article{zhang2024multi,
	title={A multi-fidelity transfer learning strategy based on multi-channel fusion},
	author={Zhang, ZiHan and Ye, Qian and Yang, DeJin and Wang, Na and Meng, GuoXiang},
	journal={Journal of Computational Physics},
	volume={506},
	pages={112952},
	year={2024},
	publisher={Elsevier}
}

@article{de2020transfer,
	title={On transfer learning of neural networks using bi-fidelity data for uncertainty propagation},
	author={De, Subhayan and Britton, Jolene and Reynolds, Matthew and Skinner, Ryan and Jansen, Kenneth and Doostan, Alireza},
	journal={International Journal for Uncertainty Quantification},
	volume={10},
	number={6},
	year={2020},
	publisher={Begel House Inc.}
}

@article{kalimullah2023probabilistic,
	title={A probabilistic framework for source localization in anisotropic composite using transfer learning based multi-fidelity physics informed neural network ({mfPINN})},
	author={Kalimullah, Nur MM and Shelke, Amit and Habib, Anowarul},
	journal={Mechanical Systems and Signal Processing},
	volume={197},
	pages={110360},
	year={2023},
	publisher={Elsevier}
}

@article{lyu2023multi,
	title={Multi-fidelity prediction of fluid flow based on transfer learning using {F}ourier neural operator},
	author={Lyu, Yanfang and Zhao, Xiaoyu and Gong, Zhiqiang and Kang, Xiao and Yao, Wen},
	journal={Physics of Fluids},
	volume={35},
	number={7},
	year={2023},
	publisher={AIP Publishing}
}

@article{meng2020composite,
	title={A composite neural network that learns from multi-fidelity data: Application to function approximation and inverse {PDE} problems},
	author={Meng, Xuhui and Karniadakis, George Em},
	journal={Journal of Computational Physics},
	volume={401},
	pages={109020},
	year={2020},
	publisher={Elsevier}
}

@article{guo2022multi,
	title={Multi-fidelity regression using artificial neural networks: Efficient approximation of parameter-dependent output quantities},
	author={Guo, Mengwu and Manzoni, Andrea and Amendt, Maurice and Conti, Paolo and Hesthaven, Jan S},
	journal={Computer Methods in Applied Mechanics and Engineering},
	volume={389},
	pages={114378},
	year={2022},
	publisher={Elsevier}
}

@article{ma2023kgml,
	title={{KGML-xDTD}: A knowledge graph--based machine learning framework for drug treatment prediction and mechanism description},
	author={Ma, Chunyu and Zhou, Zhihan and Liu, Han and Koslicki, David},
	journal={GigaScience},
	volume={12},
	pages={giad057},
	year={2023},
	publisher={Oxford University Press}
}

@article{simon2022interpretable,
	title={Interpretable machine learning prediction of drug-induced {QT} prolongation: Electronic health record analysis},
	author={Simon, Steven T and Trinkley, Katy E and Malone, Daniel C and Rosenberg, Michael Aaron},
	journal={Journal of Medical Internet Research},
	volume={24},
	number={12},
	pages={e42163},
	year={2022},
	publisher={JMIR Publications Toronto, Canada}
}

@article{yao2019prediction,
	title={Prediction of antiepileptic drug treatment outcomes of patients with newly diagnosed epilepsy by machine learning},
	author={Yao, Lijun and Cai, Mengting and Chen, Yang and Shen, Chunhong and Shi, Lei and Guo, Yi},
	journal={Epilepsy \& Behavior},
	volume={96},
	pages={92--97},
	year={2019},
	publisher={Elsevier}
}

@article{koesmahargyo2020accuracy,
	title={Accuracy of machine learning-based prediction of medication adherence in clinical research},
	author={Koesmahargyo, Vidya and Abbas, Anzar and Zhang, Li and Guan, Lei and Feng, Shaolei and Yadav, Vijay and Galatzer-Levy, Isaac R},
	journal={Psychiatry Research},
	volume={294},
	pages={113558},
	year={2020},
	publisher={Elsevier}
}

@article{su2018random,
	title={Random forests of interaction trees for estimating individualized treatment effects in randomized trials},
	author={Su, Xiaogang and Pe{\~n}a, Annette T and Liu, Lei and Levine, Richard A},
	journal={Statistics in Medicine},
	volume={37},
	number={17},
	pages={2547--2560},
	year={2018},
	publisher={Wiley Online Library}
}

@article{grzenda2021machine,
	title={Machine learning prediction of treatment outcome in late-life depression},
	author={Grzenda, Adrienne and Speier, William and Siddarth, Prabha and Pant, Anurag and Krause-Sorio, Beatrix and Narr, Katherine and Lavretsky, Helen},
	journal={Frontiers in Psychiatry},
	volume={12},
	pages={738494},
	year={2021},
	publisher={Frontiers Media SA}
}

@article{doudican2015personalization,
	title={Personalization of cancer treatment using predictive simulation},
	author={Doudican, Nicole A and Kumar, Ansu and Singh, Neeraj Kumar and Nair, Prashant R and Lala, Deepak A and Basu, Kabya and Talawdekar, Anay A and Sultana, Zeba and Tiwari, Krishna Kumar and Tyagi, Anuj and others},
	journal={Journal of Translational Medicine},
	volume={13},
	number={1},
	pages={43},
	year={2015},
	publisher={Springer}
}

@article{hoogendoorn2019broadening,
	title={Broadening the perspective of cost-effectiveness modeling in chronic obstructive pulmonary disease: A new patient-level simulation model suitable to evaluate stratified medicine},
	author={Hoogendoorn, Martine and Ramos, Isaac Corro and Baldwin, Michael and Guix, Nuria Gonzalez-Rojas and Rutten-van M{\"o}lken, Maureen PMH},
	journal={Value in Health},
	volume={22},
	number={3},
	pages={313--321},
	year={2019},
	publisher={Elsevier}
}

@article{nielsen2022robust,
	title={Robust explainability: A tutorial on gradient-based attribution methods for deep neural networks},
	author={Nielsen, Ian E and Dera, Dimah and Rasool, Ghulam and Ramachandran, Ravi P and Bouaynaya, Nidhal Carla},
	journal={IEEE Signal Processing Magazine},
	volume={39},
	number={4},
	pages={73--84},
	year={2022},
	publisher={IEEE}
}

@article{perets2024inherent,
	title={Inherent bias in electronic health records: A scoping review of sources of bias},
	author={Perets, Oriel and Stagno, Emanuela and Ben Yehuda, Eyal and McNichol, Megan and Celi, Leo Anthony and Rappoport, Nadav and Dorotic, Matilda},
	journal={ACM Transactions on Intelligent Systems and Technology},
	year={2025},
	publisher={ACM New York, NY}
}

@inproceedings{wang2024rcfr,
	title={{CE-RCFR}: Robust counterfactual regression for consensus-enabled treatment effect estimation},
	author={Wang, Fan and Chen, Chaochao and Liu, Weiming and Fan, Tianhao and Liao, Xinting and Tan, Yanchao and Qi, Lianyong and Zheng, Xiaolin},
	booktitle={Proceedings of the 30th ACM SIGKDD Conference on Knowledge Discovery and Data Mining},
	pages={3013--3023},
	year={2024}
}

@article{goetz2018personalized,
	title={Personalized medicine: Motivation, challenges, and progress},
	author={Goetz, Laura H and Schork, Nicholas J},
	journal={Fertility and Sterility},
	volume={109},
	number={6},
	pages={952--963},
	year={2018},
	publisher={Elsevier}
}

@inproceedings{WuFCL0T22,
  author       = {Jiahao Wu and
                  Wenqi Fan and
                  Jingfan Chen and
                  Shengcai Liu and
                  Qing Li and
                  Ke Tang},
  title        = {Disentangled Contrastive Learning for Social Recommendation},
  booktitle    = {Proceedings of the 31st {ACM} International Conference on Information
                  {\&} Knowledge Management},
  pages        = {4570--4574},
  year         = {2022}
}

@article{HeLHT23,
  author       = {Rui He and
                  Shengcai Liu and
                  Shan He and
                  Ke Tang},
  title        = {Multi-Domain Active Learning: Literature Review and Comparative Study},
  journal      = {IEEE Transactions on Emerging Topics in Computational Intelligence},
  volume       = {7},
  number       = {3},
  pages        = {791--804},
  year         = {2023}
}

@ARTICLE{Wu2024TKDE,
  author={Wu, Jiahao and Fan, Wenqi and Chen, Jingfan and Liu, Shengcai and Liu, Qijiong and He, Rui and Li, Qing and Tang, Ke},
  journal={IEEE Transactions on Knowledge and Data Engineering}, 
  title={Condensing Pre-Augmented Recommendation Data via Lightweight Policy Gradient Estimation}, 
  year={2025},
  volume={37},
  number={1},
  pages={162-173}
}

@inproceedings{Wu2025WWW,
author = {Wu, Jiahao and Liu, Qijiong and Hu, Hengchang and Fan, Wenqi and Liu, Shengcai and Li, Qing and Wu, Xiao-Ming and Tang, Ke},
title = {Leveraging {ChatGPT} to Empower Training-free Dataset Condensation for Content-based Recommendation},
year = {2025},
isbn = {9798400713316},
publisher = {{ACM}},
booktitle = {Companion Proceedings of the ACM on Web Conference 2025},
pages = {1402–1406}
}

@ARTICLE{11481606,
	author={Du, Wenbo and Han, Lingling and Xiong, Ying and Zhang, Ling and Li, Biyue and Lv, Yisheng and Guo, Tong},
	journal={IEEE Transactions on Intelligent Transportation Systems}, 
	title={Airport Passenger Flow Forecasting via Deformable Temporal–Spectral Transformer Approach}, 
	year={2026},
	volume={},
	number={},
	pages={1-15},
	doi={10.1109/TITS.2026.3681954}}

@ARTICLE{11601056,
		author={Liu, Shengcai and Wang, Zhiyuan and Ong, Yew-Soon and Yao, Xin and Tang, Ke},
		journal={IEEE Transactions on Pattern Analysis and Machine Intelligence}, 
		title={{MEGO}: Learning Mixture-of-Experts for General-Purpose Binary Optimization}, 
		year={2026},
		volume={},
		number={},
		pages={1-18},
		doi={10.1109/TPAMI.2026.3711708}}

@inproceedings{belghazi2018mutual,
	title={Mutual information neural estimation},
	author={Belghazi, Mohamed Ishmael and Baratin, Aristide and Rajeshwar, Sai and Ozair, Sherjil and Bengio, Yoshua and Courville, Aaron and Hjelm, Devon},
	booktitle={International conference on machine learning},
	pages={531--540},
	year={2018},
	organization={PMLR}
}

\begin{IEEEbiography}[{\includegraphics[width=1in,height=1.25in,clip,keepaspectratio]{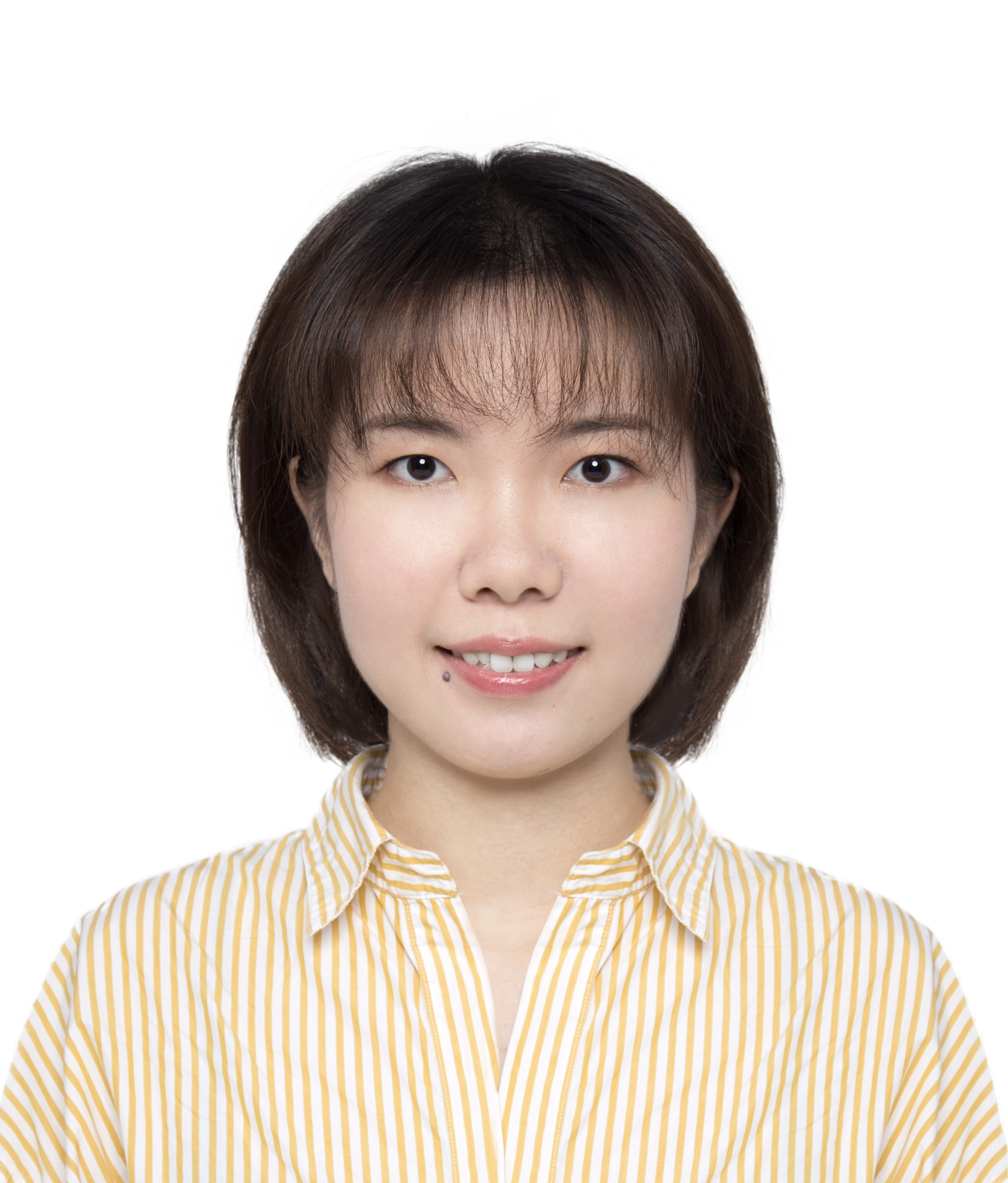}}]{Wenjie Chen}
	received the B.S. degree in automation from the Huazhong University of Science and Engineering, Wuhan, China, in 2016,  and Ph.D. degree in systems engineering and engineering management from the City University of Hong Kong, Hong Kong, in 2020. She is currently an Assistant Professor with the School of Information Management, Central China Normal University, Wuhan, China. Her research is devoted to simulation optimization, machine learning, and real-world applications such as healthcare management and social network analysis.
\end{IEEEbiography}

\begin{IEEEbiography}[{\includegraphics[width=1in,height=1.25in,clip,keepaspectratio]{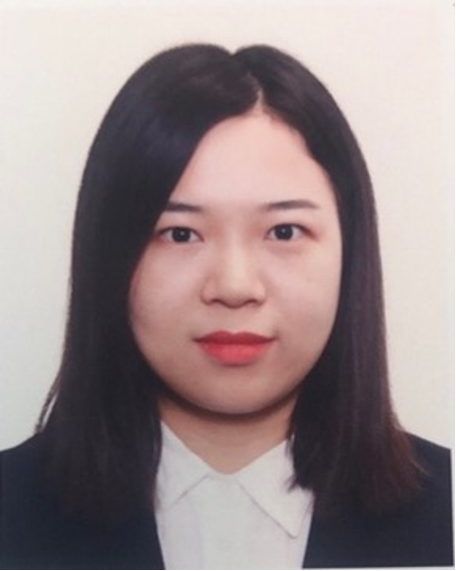}}]{Li Zhuang}
	received the B.Eng. degree in automation from the Huazhong University of Science and Technology, Wuhan, China, in 2016, and the Ph.D. degree in data science from the City University of Hong Kong, Hong Kong, in 2020. She is currently an Assistant Professor with the School of Cyber Science and Engineering, Southeast University, Nanjing, China. Her research interests focus on data mining, machine learning, and computer vision with applications on intelligent transportation and renewable energy.
\end{IEEEbiography}

\begin{IEEEbiography}[{\includegraphics[width=1in,height=1.25in,clip,keepaspectratio]{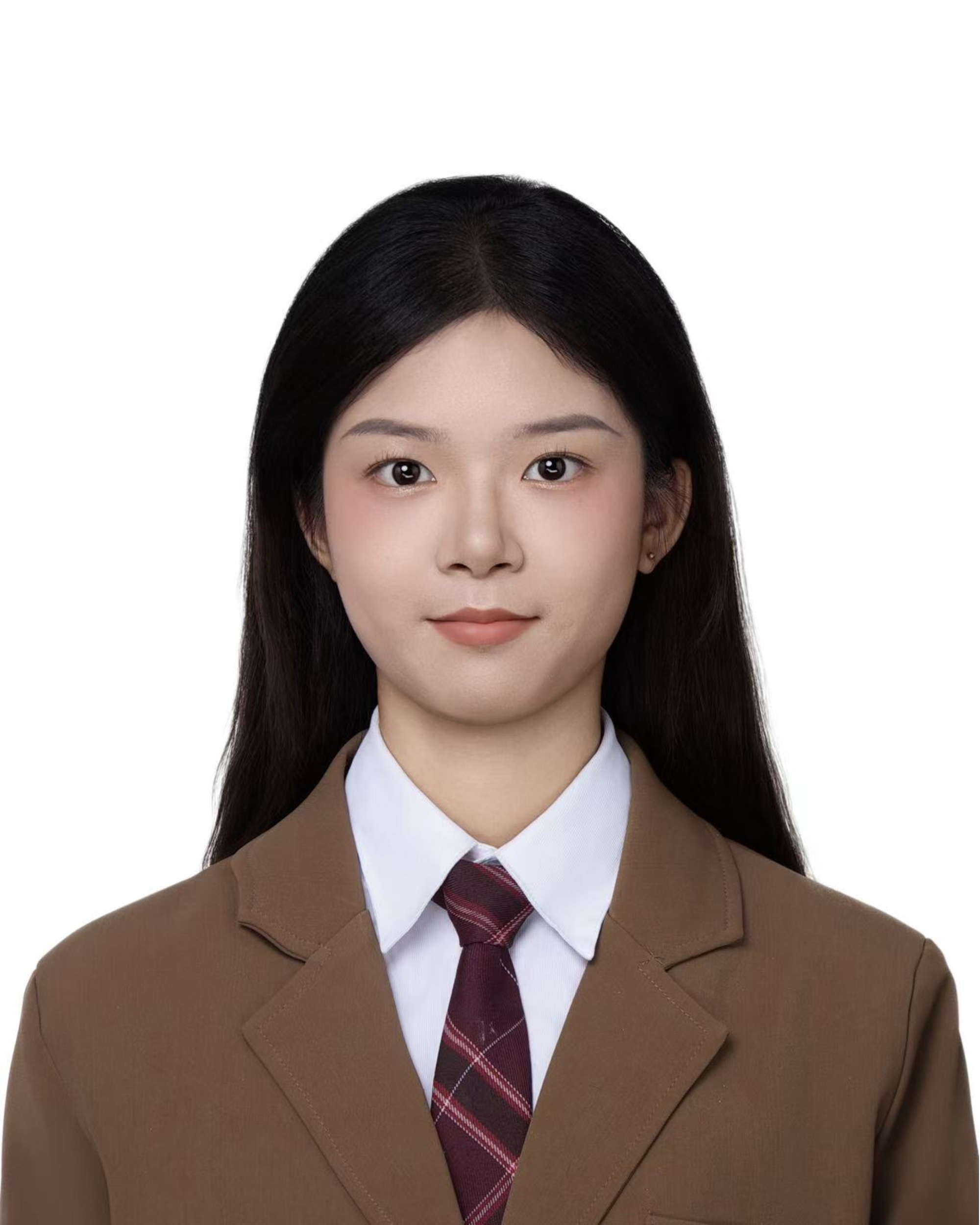}}]{Ziying Luo}
	(Student Member, IEEE) received the B.Eng. degree in big data management and application from Central China Normal University (CCNU), Wuhan, China, in 2025. She is currently pursuing the M.S. degree  in the Department of Computer Science and Engineering at Southern University of Science and Technology (SUSTech), Shenzhen, China. Her research interests include machine learning and agent safety.
\end{IEEEbiography}

\begin{IEEEbiography}[{\includegraphics[width=1in,height=1.25in,clip,keepaspectratio]{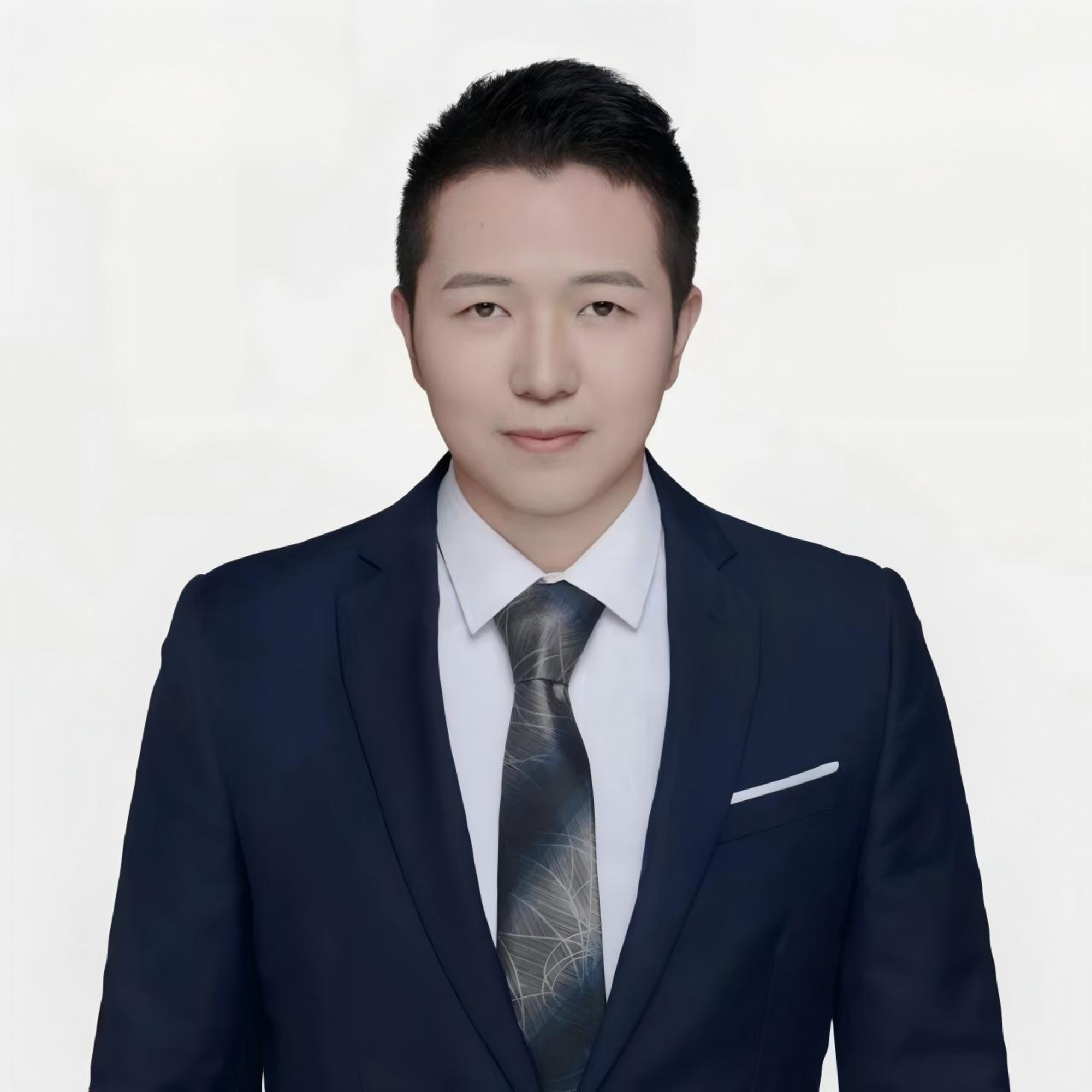}}]{Yu Liu}
	received the Ph.D. degree in software engineering from the School of Computer Science, Wuhan University, Wuhan, China, in 2024. He is currently with the School of Information Management, Central China Normal University, Wuhan, China. His research interests include data mining, large language models, harness engineering for LLM-based agents, information and network security, causal discovery and inference, services computing, software engineering, and multimedia technology.
\end{IEEEbiography}

\begin{IEEEbiography}[{\includegraphics[width=1in,height=1.25in,clip,keepaspectratio]{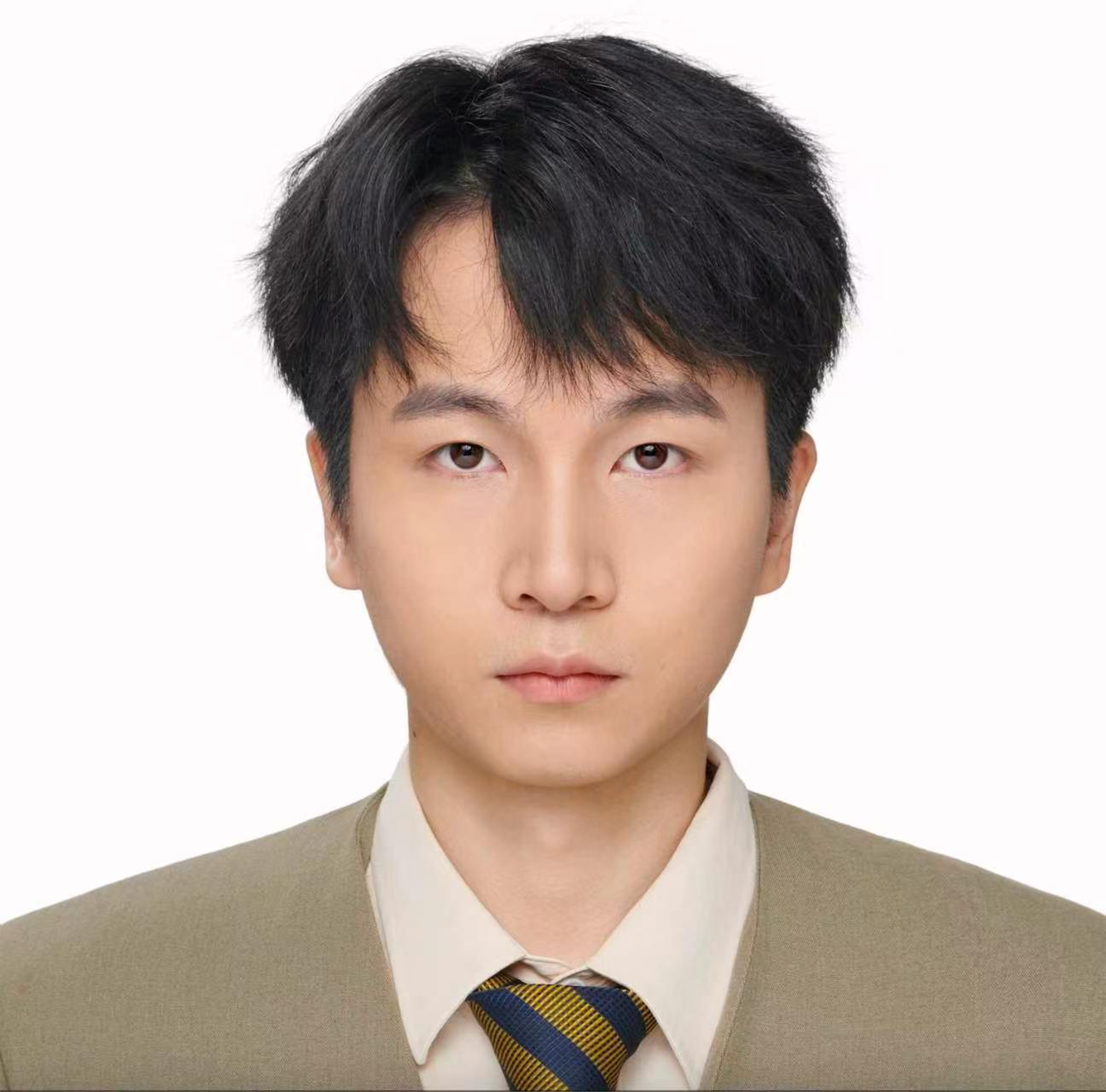}}]{Jiahao Wu}
	received the B.E. degree in computer science from the University of Science and Technology of China in 2020 and the Ph.D. degree in computing from The Hong Kong Polytechnic University in 2025. He is currently a Postdoctoral Fellow with the Department of Computing, The Hong Kong Polytechnic University. His research interests include data-efficient AI, reinforcement learning for large reasoning models, recommender systems, dataset condensation, graph learning, and LLM safety. His work has appeared in IEEE TKDE, ICML, ICDE, The Web Conference, and CIKM. He has also served as a reviewer for major conferences and journals, including NeurIPS, ICML, EMNLP, IEEE TKDE, ACM TOIS, and ACM TKDD.
\end{IEEEbiography}

\begin{IEEEbiography}[{\includegraphics[width=1in,height=1.25in,clip,keepaspectratio]{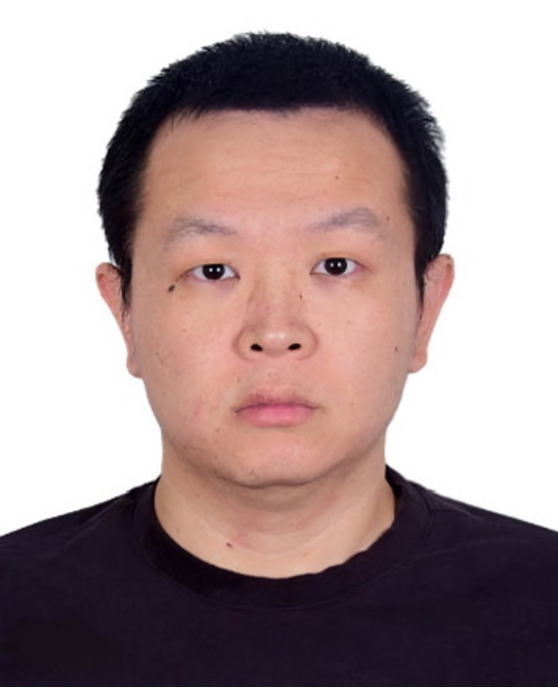}}]{Shengcai Liu}
	(Member, IEEE) received the B.Sc. degree in 2014 and the Ph.D. degree in 2020 from the University of Science and Technology of China, Hefei, China. He is currently an Assistant Professor with the Department of Computer Science and Engineering, Southern University of Science and Technology, Shenzhen, China. He is mainly interested in Learning to Optimize and has published more than 30 papers in top-tier refereed international conferences and journals. He is also an associate editor of the IEEE Transactions on Emerging Topics in Computational Intelligence.
\end{IEEEbiography}

\end{document}